\documentclass{article}

\usepackage[final]{ap_2017}

\usepackage[utf8]{inputenc} 
\usepackage[T1]{fontenc}    
\usepackage{hyperref}       
\usepackage{url}            
\usepackage{booktabs}       
\usepackage{amsmath}
\usepackage{amsfonts}       
\usepackage{nicefrac}       
\usepackage{microtype}      
\usepackage{natbib}
\usepackage{graphicx}
\usepackage[linesnumbered,ruled,vlined]{algorithm2e}
\usepackage{paralist}
\usepackage{tabularx}
\usepackage{float}

\usepackage{blindtext}
\usepackage{scrextend}
\addtokomafont{labelinglabel}{\sffamily}
\usepackage{minibox}
\usepackage{amsthm}

\usepackage{titlesec}
\titleformat{\paragraph}
{\normalfont\normalsize\bfseries}{\theparagraph}{1em}{}
\titlespacing*{\paragraph}
{0pt}{3.25ex plus 1ex minus .2ex}{1.5ex plus .2ex}

\DeclareMathOperator*{\argmax}{arg\,max}
\newcommand{\defeq}{\stackrel{\text{def}}{=}}

\title{A Deep Reinforcement Learning Chatbot}
\author{
  \normalfont \textbf{Iulian V. Serban}, \textbf{Chinnadhurai Sankar}, \textbf{Mathieu Germain}, \textbf{Saizheng Zhang}, \textbf{Zhouhan Lin}, \\ \textbf{Sandeep Subramanian},
  \textbf{Taesup Kim}, \textbf{Michael Pieper}, \textbf{Sarath Chandar}, \textbf{Nan Rosemary Ke}, \\ \textbf{Sai Rajeshwar}, \textbf{Alexandre de Brebisson},
  \textbf{Jose M. R. Sotelo}, \textbf{Dendi Suhubdy}, \\ \textbf{Vincent Michalski}, \textbf{Alexandre Nguyen}, \textbf{Joelle Pineau}$^{1,2}$ and \textbf{Yoshua Bengio}$^2$ \\
  Montreal Institute for Learning Algorithms, Montreal, Quebec, Canada
}

\begin{document}

\maketitle

\begin{abstract}
We present MILABOT: a deep reinforcement learning chatbot developed by the Montreal Institute for Learning Algorithms (MILA) for the Amazon Alexa Prize competition.
MILABOT is capable of conversing with humans on popular small talk topics through both speech and text.
The system consists of an ensemble of natural language generation and retrieval models, including template-based models, bag-of-words models, sequence-to-sequence neural network and latent variable neural network models.
By applying reinforcement learning to crowdsourced data and real-world user interactions, the system has been trained to select an appropriate response from the models in its ensemble.
The system has been evaluated through A/B testing with real-world users, where it performed significantly better than many competing systems.
Due to its machine learning architecture, the system is likely to improve with additional data.
\end{abstract}

\stepcounter{footnote}
\footnotetext{School of Computer Science, McGill University.}

\stepcounter{footnote}
\footnotetext{CIFAR Fellow.}

\section{Introduction}
Dialogue systems and conversational agents - including chatbots, personal assistants and voice-control interfaces - are becoming ubiquitous in modern society.
Examples of these include personal assistants on mobile devices, technical support help over telephone lines, as well as online bots selling anything from fashion clothes and cosmetics to legal advice and self-help therapy.
However, building intelligent conversational agents remains a major unsolved problem in artificial intelligence research.

In 2016, Amazon.com Inc proposed an international university competition with the goal of building a socialbot: a spoken conversational agent capable of conversing coherently and engagingly with humans on popular topics, such as entertainment, fashion, politics, sports, and technology.
The socialbot converses through natural language speech through Amazon's Echo device \citep{stone2014echo}.
This article describes the models, experiments and final system (MILABOT) developed by our team at University of Montreal.\footnote{Our team is called MILA Team, where MILA stands for the Montreal Institute for Learning Algorithms.}
Our main motivation for participating has been to help advance artificial intelligence research.
To this end, the competition has provided a special opportunity for training and testing state-of-the-art machine learning algorithms with real users (also known as \textit{machine learning in the wild}) in a relatively unconstrained setting.
The ability to experiment with real users is unique in the artificial intelligence community, where the vast majority of work consists of experiments on fixed datasets (e.g.\@ labeled datasets) and software simulations (e.g.\@ game engines).
In addition, the computational resources, technical support and financial support provided by Amazon has helped scale up our system and test the limits of state-of-the-art machine learning methods.
Among other things, this support has enabled us to crowdsource ~$200,000$ labels on Amazon Mechanical Turk and to maintain over 32 dedicated Tesla K80 GPUs for running our live system.

Our socialbot is based on a large-scale ensemble system leveraging deep learning and reinforcement learning.
We develop a new set of deep learning models for natural language retrieval and generation --- including recurrent neural networks, sequence-to-sequence models and latent variable models --- and evaluate them in the context of the competition.
These models are combined into an ensemble, which generates a candidate set of dialogue responses.
Further, we apply reinforcement learning --- including value function and policy gradient methods --- to train the system to select an appropriate response from the models in its ensemble.
In particular, we propose a novel reinforcement learning procedure, based on estimating a Markov decision process.
Training is carried out on crowdsourced data and on interactions recorded between real-world users and a preliminary version of the system.
The trained systems yield substantial improvements in A/B testing experiments with real-world users. 


In the competition semi-finals, our best performing system reached an average user score of $3.15$ on a scale $1-5$, with a minimal number of hand-crafted states and rules and without engaging in \textit{non-conversational activities} (such as playing games or taking quizzes).\footnote{Throughout the semi-finals we carried out several A/B testing experiments to evaluate different variants of our system (see Section \ref{sec:ab_testing_experiments}). The score $3.15$ is based on the best performing system in the period between July 29th and August 6th, 2017. The score is not based on the leaderboard, which averages the scores of all the variants of our system (including a supervised learning system and a heuristic baseline system).}
The performance of this best system is substantially better than the average of all the teams in the competition semi-finals.
Further, the same system averaged a high $14.5-16.0$ turns per dialogue, which is also significantly higher than the average of all the teams in the competition semi-finals, as well as the finalist teams.
This improvement in back-and-forth exchanges between the user and system suggests that our system is likely to be the most engaging system among all systems in the competition.
Finally, the system is bound to improve with additional data, as nearly all system components are learnable.

\section{System Overview}\label{sec:dialogue_system_overview}

Early work on dialogue systems~\citep{weizenbaum1966eliza,BBS:2533884,aust1995philips,mcglashan1992dialogue,simpson1993black} were based mainly on states and rules hand-crafted by human experts.
Modern dialogue systems typically follow a hybrid architecture, combining hand-crafted states and rules with statistical machine learning algorithms~\citep{suendermann2015halef,jurvcivcek2014alex,bohus2007olympus,williams2011empirical}. 
Due to the complexity of human language, however, it will probably never be possible to enumerate states and rules required for building a socialbot capable of conversing with humans on open-domain, popular topics.
In contrast to such rule-based systems, our core approach is built entirely on statistical machine learning.
We believe that this is the most plausible path to artificially intelligent conversational agents.
The system architecture we propose aims to make as few assumptions as possible about the process of understanding and generating natural human language.
As such, the system utilizes only a small number of hand-crafted states and rules.
However, every system component has been designed to be optimized (trained) using machine learning algorithms. 
These system components will be trained first independently on massive datasets and then jointly on real-world user interactions.
This way, the system will learn all relevant states and rules for conducting open-domain conversations implicitly.
Given an adequate amount of examples, such a system should outperform systems based on hand-crafted states and rules.
Further, the system will continue to improve in perpetuity with additional data.


\begin{figure}[ht]
  \centering
  \includegraphics[scale=0.25]{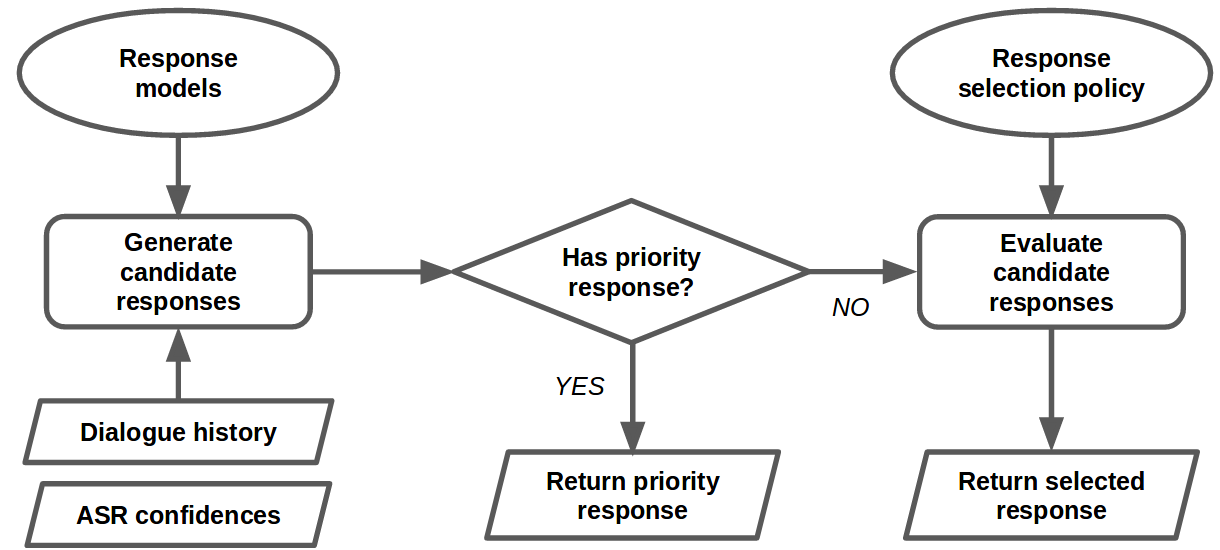}
  \caption{Dialogue manager control flow.}
  \label{fig:dialogue_system}
\end{figure}

Our system architecture is inspired by the success of ensemble-based machine learning systems.
These systems consist of many independent sub-models combined intelligently together.
Examples of such ensemble systems include the winner of the Netflix Prize~\citep{koren2009matrix}, utilizing hundreds of machine learning models to predict user movie preferences, and IBM Watson~\citep{ferrucci2010building}, the first machine learning system to win the quiz game Jeopardy! in 2011.
More recently, Google observed substantial improvements building an ensemble-based neural machine translation system~\citep{wu2016google}.

Our system consists of an ensemble of response models.
The response models take as input a dialogue and output a response in natural language text.
In addition, the response models may also output one or several scalar values, indicating their internal confidence.
As will be explained later, the response models have been engineered to generate responses on a diverse set of topics using a variety of strategies.

The \textit{dialogue manager} is responsible for combining the response models together.
As input, the dialogue manager expects to be given a dialogue history (i.e.\@ all utterances recorded in the dialogue so far, including the current user utterance) and confidence values of the automatic speech recognition system (ASR confidences).
To generate a response, the dialogue manager follows a three-step procedure.
First, it uses all response models to generate a set of candidate responses.
Second, if there exists a \textit{priority} response in the set of candidate responses (i.e.\@ a response which takes precedence over other responses), this response will be returned by the system.\footnote{An ordering of the models decides which response to return in case there are multiple \textit{priority} responses.}
For example, for the question \textit{"What is your name?"}, the response \textit{"I am an Alexa Prize socialbot"} is a priority response.
Third, if there are no \textit{priority} responses, the response is selected by the \textit{model selection policy}.
For example, the \textit{model selection policy} may select a response by scoring all candidate responses and picking the highest-scored response.
The overall process is illustrated in Figure \ref{fig:dialogue_system}.

When the ASR confidences are below a certain threshold, the system requests the user to repeat their last utterance.
Otherwise, the system does not utilize the ASR confidences.
In particular, neither the response models nor the \textit{model selection policy} make use of the ASR confidences.
Nevertheless, the ASR system is far from perfect.
Therefore, it is very plausible that the system could be improved significantly by conditioning the response models and \textit{model selection policy} on the ASR confidences.

\section{Response Models}
There are 22 response models in the system, including retrieval-based neural networks, generation-based neural networks, knowledge base question answering systems and template-based systems.
Examples of candidate model responses are shown in Table \ref{table:candidate_response-examples}.
This section aims to summarize these response models.
First, the template-based models are described.

\newcolumntype{L}[1]{>{\raggedright\arraybackslash}p{#1}}

\begin{table*}[ht]
 \caption{Example dialogues and corresponding candidate responses generated by response models. The response of the final system is marked in bold.}
 \label{table:candidate_response-examples}
 \small
 \centering
 \begin{tabular}{L{58mm}L{72mm}} \toprule
 \textbf{Dialogue} & \textbf{Dialogue} \\ \hline
    SYSTEM: Hi. This is an Alexa Prize socialbot. \newline USER: Hi \newline SYSTEM: Do you have pets? \newline USER: Yes, I have two rabbits. & ... \newline USER: Let's talk about movies. \newline SYSTEM: What's your favorite movie? \newline USER: star wars movie \\ \\
     \textbf{Candidate Responses} & \textbf{Candidate Responses} \\ \hline
    BoWEscapePlan: I don't have an answer for this. \newline BoWFactGenerator: Did you know that male rabbits are called bucks, females are does? \newline GRUQuestionGenerator: How about the American family of it? \newline Initiatorbot: Do you believe in love at first sight? \newline LSTMClassifierMSMarco: Mar 23, 2017  Two Parts: Learning What to Feed Your Rabbit Learning How to ... \newline DualEncoderRedditNews: bye doggie \newline DualEncoderRedditPolitics: Who's edger u shboys? \newline \textbf{Alicebot: Hurrah! Two is a good number of rabbits.} \newline Elizabot: OK, but can you elaborate a bit? \newline  VHREDRedditMovies: Hmm. Nope, never heard of it. \newline VHREDSubtitles: Would you like to see them? \newline HelpResponse: I would really like to talk about news, politics or movies. & BoWEscapePlan: I don't have an answer for this. \newline BoWFactGenerator: Here's an interesting fact. The international telephone dialing code for Antarctica is 672. \newline GRUQuestionGenerator: how about your New York City? \newline Initiatorbot: How was your day? \newline LSTMClassifierMSMarco: A third Anthology film will be released in 2020. \newline DualEncoderRedditNews: Now whisper it to me , one more time bby. \newline DualEncoderRedditPolitics: hahaha on mobile this ain't even close. I wish I could show you but this looks awful I'm sorry man. \newline Alicebot: What did you like about the robots in that movie? \newline Elizabot: How does that make you feel? \newline \textbf{Evi: Star Wars movie a movie in the Star Wars series.}   \newline VHREDRedditMovies: Oh please. Please. Pleeeease. Let this happen. \newline VHREDSubtitles: What? \newline HelpResponse: I would really like to talk about news, politics or movies. \\
 \bottomrule
 \end{tabular}
\end{table*}

\subsection{Template-based Models}
We start by describing the template-based response models in the system.

\textbf{Alicebot}: \emph{Alicebot} uses a set of AIML (artificial intelligence markup language) templates to produce a response given the dialogue history and user utterance~\citep{wallace2009anatomy,shawar2007chatbots}.
We use the freely available Alice kernel available at \url{www.alicebot.org}.
By default all templates generate non-priority responses, so we configure templates related to the socialbot's name, age and location to output priority responses.
We modify a few templates further to make them consistent with the challenge (e.g.\@ to avoid obscene language and to encourage the user to discuss certain topics, such as news, politics and movies).
The majority of templates remain unchanged.

The \emph{Alicebot} model also outputs a scalar confidence score.
Since the AIML templates repeat the user's input utterance, they are not always correct sentences.
Therefore, we use a string-based rules to determine if the response constitutes a correct sentence.
If the response is correct sentence, it returns a high confidence and otherwise it returns a low confidence score.
This process is illustrated in Algorithm \ref{algorithm:alicebot}.
\begin{algorithm}[H]
\caption{Alicebot} \label{algorithm:alicebot}
\textbf{input}: dialogue history \\
response $\gets$ {apply AIML templates to dialogue history} \\
\If{response is correct sentence}{
\If{response is given priority}{confidence $\gets$ 1.0}
\Else{confidence $\gets$ 0.5}
}
\Else{confidence $\gets$ 0.0}
\textbf{output}: response, priority, confidence
\end{algorithm}

\textbf{Elizabot}
Similar to \emph{Alicebot}, the \emph{Elizabot} model performs string matching to select an answer from a set of templates.
The model is based on the famous Eliza system, designed to mimic a Rogerian psychotherapist.
~\citep{weizenbaum1966eliza}.\footnote{We use the implementation available at: \url{https://gist.github.com/bebraw/273706}.}
Therefore, in contrast with \emph{Alicebot}, most of \emph{Elizabot}'s responses are personal questions which are meant to engage the user to continue the conversation.

\newpage
Here are two example templates:
\begin{enumerate}
\item \textit{"I am (.*)" $\to$ "Did you come to me because you are ..."}
\item \textit{"What (.*)" $\to$  "Why do you ask?"}
\end{enumerate}

The ellipses mark the parts of the response sentence which will be replaced with text from the user's utterance.
The model detects the appropriate template and selects the corresponding response (if there are multiple templates, then a template is selected at random).
The model then runs the template response through a set of \textit{reflections} to better format the string for a response (e.g.\@ \textit{"I'd" $\to$ "you would"}, \textit{"your" $\to$ "my"}).


\begin{algorithm}[H]
\caption{Initiatorbot} \label{algorithm:initiatorbot}
\textbf{input}: dialogue history \\
\uIf{Initiatorbot was triggered in one of last two turns}{return ""}
\uElseIf{user did not give a greeting}{return a non-priority response with a random initiator phrase}
\Else{return a priority response with a random initiator phrase}
\end{algorithm}

\textbf{Initiatorbot}
The \emph{Initiatorbot} model acts as a \textit{conversation starter}:
it asks the user an open-ended question to get the conversation started and increase the engagement of the user.
We wrote 40 question phrases for the \emph{Initiatorbot}.
Examples of phrases include \textit{"What did you do today?"}, \textit{"Do you have pets?"} and \textit{"What kind of news stories interest you the most?"}.
As a special case, the model can also start the conversation by stating an interesting fact.
In this case, the initiator phrase is \textit{"Did you know that <fact>?"}, where \textit{fact} is replaced by a statement.
The set of facts is the same as used by the \emph{BoWFactGenerator} model, described later.

Before returning a response, \emph{Initiatorbot} first checks that it hasn't already been triggered in the last two turns of the conversation.
If the user gives a greeting (e.g.\@ \textit{"hi"}), then \emph{Initiatorbot} will return a response with priority.
This is important because we observed that greetings often indicate the beginning of a conversation, where the user does not have a particular topic they would like to talk about.
By asking a question, the system takes the \textit{initiative} (i.e.\@ control of the dialogue).
The procedure is detailed in Algorithm \ref{algorithm:initiatorbot}.

\textbf{Storybot}
The \emph{Storybot} model outputs a short fiction story at the request of the user.
We implemented this model as we observed that many users were asking the socialbot to tell stories.\footnote{Requests for telling stories is possibly a side-effect of user's interacting with bots from other teams, which often emphasized \textit{non-conversational activities}, such as telling stories and playing quizzes and word games.}
\emph{Storybot} determines if the user requested a story by checking if there was both a request word (e.g.\@ \textit{say}, \textit{tell}.) and story-type word in the utterance (e.g.\@ \textit{story}, \textit{tale}).
The response states the story's title and author followed by the story body.
For example, one set of responses from this model follows the pattern \textit{"Alright, let me tell you the story <story\_title> <story\_body> by <story\_author>"} where <story\_title> is the title of the story, <story\_body> is the main text and <story\_author> is the name of the story's author.
The stories were scraped from the website: \url{www.english-for-students.com}.


An example story is:

\minibox[frame]{
\textbf{** The Ant and The Grasshopper **}\\
The ants worked hard in summer.
They sorted food for winter.\\
At that time, a grasshopper remained idle.
When winter came, the ants had enough to eat.\\
But, the grasshopper had nothing to eat.
He had to starve.\\
He went to the ants and begged for foods.
The ants asked in return,
"What did you do in summer?"\\
He replied, "I idled away my time during summer". \\
The ant replied, "Then you must starve in winter." MORAL: Never be idle.}

The \emph{Storybot} is the only component in the system performing a \textit{non-conversational activity}.
It is triggered only when a user specifically asks for a story, and in that case its response is a priority response.
Otherwise, the \emph{Storybot} response model is never triggered.
Further, the rest of the system will not encourage the user to request stories.

\subsection{Knowledge Base-based Question Answering}

\textbf{Evibot}
The \emph{Evibot} response model forwards the user's utterance to Amazon's question-answering web-service \textit{Evi}: \url{www.evi.com}.
\textit{Evi} was designed primarily to handle factual questions.
Therefore, \textit{Evibot} returns a priority response for direct questions, defined as user utterances containing a wh-word (e.g.\@ \textit{"who"}, \textit{"what"}), and otherwise returns a non-priority or, possibly, an empty response.
If the query is a direct question and contains non-stop words, \emph{Evibot} will follow a three step procedure to generate its response.
First, \emph{Evibot} forwards a query to \url{www.evi.com} containing the whole user utterance, and returns the resulting answer if its valid.
If that fails, \emph{Evibot} applies NLTK's named entity processor~\citep{BirdKleinLoper09} to the query to find subqueries with named entities.
For each subphrase that contains a named entity, \emph{Evibot} forwards queries to \url{www.evi.com}, and returns the result upon a valid response.
Finally, if the previous two steps fail, \emph{Evibot} forwards queries for every subquery without named entities, and returns either a valid response or an empty response.
The procedure is detailed in Algorithm \ref{algorithm:evibot}.
\begin{algorithm}[H]
\caption{Evibot} \label{algorithm:evibot}
\textbf{input}: dialogue history \\
query $\gets$ last user utterance \\
has-wh-words $\gets$ true if utterance contains a wh-word, otherwise false \\
has-only-stop-words $\gets$ true if utterance only has stop words, otherwise false \\
\If{has-only-stop-words and not has-wh-words}{return ""}
evi-response $\gets$ send query to \url{www.evi.com} \\
priority $\gets$ true if has-wh-words and evi-response is valid, otherwise false  \\
\If{evi-response is valid}{return evi-response, priority}
\uElseIf{has-wh-words}{
priority $\gets$ has-wh-words \\
subentities $\gets$ entities extracted from query using NLTK's named entity processor \\
subphrases $\gets$ list of subphrases with entities \\
\For{subphrase in subphrases}{
evi-response $\gets$ send subphrase to \url{www.evi.com} \\
\If{evi-response is valid}{
return evi-response, priority}
}
subphrases $\gets$ list of all subphrases \\
\For{subphrase in subphrases}{
evi-response $\gets$ send subphrase to \url{www.evi.com} \\
\If{evi-response is valid}{
return evi-response, priority}
}
}
\Else {return ""}
\end{algorithm}

\textbf{BoWMovies}
The \emph{BoWMovies} model is a template-based response model, which handles questions in the movie domain.
The model has a list of entity names and tags (e.g.\@ \textit{movie plot} and \textit{release year}).
The model searches the user's utterance for known entities and tags.
Entities are identified by string matching.
This is done in a cascading order, by giving first preference to movie title matches, then actor name matches, and finally director name matches.
Tags are also identified by string matching.
However, if exact string matching fails for tags, then identification is performed by word embedding similarity.
If both an entity and a tag are present, the agent will dispatch an API call to one of several data sources to retrieve the data item for the selected query type.
The agent is limited by the data available in the APIs to which it has access.
The model's responses follow predefined templates.

Movie titles, actor names, and director names are extracted from the Internet Movie Database (IMDB).
Movie descriptions are taken from Google Knowledge Graph’s API. Other movie title queries are directed to the Open Movie Database (OMDB).\footnote{See \url{www.omdbapi.com}. This should not be confused with IMDB.}
For actor and director queries,  the Wikiedata API is used.
First, a search for actor and director names is done on a Wikidata JSON dump.


As described earlier, the model uses word embeddings to match tags.
These word embeddings are trained using Word2Vec on movie plot summaries and actor biographies extracted from the IMDB database~\citep{mikolov2013distributedbetter}. 


\begin{algorithm}[H]
\caption{BoWMovies - ComputeResponse}
\textbf{input}: dialogue history \\
entity $\gets$ entity contained both in last user utterance and list of movie titles, actors or directors \\
\If{no entity}{entity $\gets$ entity contained in previous user utterances and movie titles, actors or directors}
\If{no entity}{return ""}
\uIf{entity is a movie title}{response $\gets$ ComputeEntityResponse(entity, movie title)}
\uElseIf{entity is an actor name}{response $\gets$ ComputeEntityResponse(entity, actor name)}
\uElseIf{entity is an director name}{response $\gets$ ComputeEntityResponse(entity, director name)}
return response
\end{algorithm}

\begin{algorithm}[H]
\caption{BoWMovies - ComputeEntityResponse}
\textbf{input}: entity and entity type \\
tag $\gets$ string matching tag, where tag is valid for entity type (movie title, actor name, director name) \\
\If{no tag }{
tag $\gets$ word embedding matching tag, where tag is a single word and valid for the entity type (movie title, actor name, director name)}
\If{no tag}{
tag $\gets$ word embedding matching tag, where tag is multiple words and valid for the entity type (movie title, actor name, director name)}
\If{no tag}{return ""}
api-response $\gets$ call external API with query (entity, tag).\\
response $\gets$ template with api-response inserted \\
return response
\end{algorithm}

\subsection{Retrieval-based Neural Networks}
\textbf{VHRED models}:
The system contains several VHRED models, sequence-to-sequence models with Gaussian latent variables trained as variational auto-encoders~\citep{serban2016hierarchical,kingma2013auto,rezende2014stochastic}.
The models are trained using the same procedure as~\citet{serban2016hierarchical}.
A comparison between VHRED and other generative sequence-to-sequence models is provided by~\citet{serban2016generative}.
The trained VHRED models generate candidate responses as follows.
First, a set of $K$ model responses are retrieved from a dataset using cosine similarity between the current dialogue history and the dialogue history in the dataset based on bag-of-words TF-IDF Glove word embeddings~\citep{pennington2014glove}.\footnote{We use the Glove embeddings trained on Wikipedia 2014 + Gigaword 5: \url{https://nlp.stanford.edu/projects/glove/}.}
An approximation of the log-likelihood for each of the 20 responses is computed by VHRED, and the response with the highest log-likelihood is returned.
The system has 4 VHRED models based on datasets scraped from Reddit, one VHRED model based on news articles and one VHRED model based on movie subtitles:
\begin{compactitem}
\item \emph{VHREDRedditPolitics} trained on \url{https://www.reddit.com/r/politics} and extracting responses from all Reddit datasets with $K=10$,
\item \emph{VHREDRedditNews} trained on Reddit \url{https://www.reddit.com/r/news} and extracting responses from all Reddit datasets with $K=20$,
\item \emph{VHREDRedditSports} trained on Reddit \url{https://www.reddit.com/r/sports} and extracting responses from all Reddit datasets with $K=20$, 
\item \emph{VHREDRedditMovies} trained on Reddit \url{https://www.reddit.com/r/movies} and extracting responses from all Reddit datasets with $K=20$,
\item \emph{VHREDWashingtonPost}\footnote{For \emph{VHREDWashingtonPost}, the $K$ responses are extracted based on the cosine similarity between the current dialogue and the news article keywords. $K$ varies depending on the number of user comments within a set of news articles above a certain cosine similarity threshold.} trained on Reddit \url{https://www.reddit.com/r/politics} and extracting responses from user comments to WashingtonPost news articles, and
\item \emph{VHREDSubtitles}\footnote{For \emph{VHREDSubtitles}, cosine similarity is computed based on one-hot vectors for each word.} using the movie subtitles dataset SubTle~\citep{ameixa2014luke} with $K=10$.
\end{compactitem}

In particular, \emph{VHREDRedditPolitics} and \emph{VHREDWashingtonPost} use a different retrieval procedure.
These two models use a logistic regression model to score the responses instead of the approximate log-likelihood.
The logistic regression model is trained on a set of ~$7500$ Reddit threads and candidate responses annotated by Amazon Mechanical Turk workers on a Likert-type scale $1-5$.
The candidate responses are selected from other Reddit threads according to cosine similarity w.r.t.\@ Glove word embeddings.
The label collection and training procedure for the logistic regression model are similar to the procedures described in Section \ref{sec:model_selection_policy}.
For each response, the logistic regression model takes as input the VHRED log-likelihood score, as well as several other input features, and outputs a scalar-valued score.
Even though the logistic regression model did improve the appropriateness of responses selected for Reddit threads, \emph{VHREDRedditPolitics} is used extremely rarely in the final system (see Section \ref{sec:model_selection_policy}).
This suggests that training a model to rerank responses based on labeled Reddit threads and responses cannot help improve performance.

\textbf{SkipThought Vector Models}:
The system contains a SkipThought Vector model~\citep{kiros2015skip} trained on the BookCorpus dataset~\citep{zhu2015aligning} and on the SemEval 2014 Task 1~\citep{marelli2014semeval}.
The model was trained using the same procedure as~\citet{kiros2015skip} and is called \emph{SkipThoughtBooks}.

\emph{SkipThoughtBooks} ensures that the system complies with the Amazon Alexa Prize competition rules.
One rule, introduced early in the competition, is that socialbots were not supposed to state their own opinions related to political or religious topics.
If a user wishes to discuss such topics, the socialbots should proceed by asking questions or stating facts.
\emph{SkipThoughtBooks} also handles idiosyncratic issues particular to the Alexa platform.
For example, many users did not understand the purpose of a socialbot and asked our socialbot to play music. In this case, the system should instruct the user to exit the \textit{socialbot application} and then play music.

\emph{SkipThoughtBooks} follows a two-step procedure to generate its response.
The first step compares the user's last utterance to a set of trigger phrases.
If a match is found, the model returns a corresponding priority response.\footnote{Trigger phrases may have multiple responses. In this case, a response is selected at random.}
For example, if the user says \textit{"What do you think about Donald trump?"}, the model will return a priority response, such as \textit{"Sometimes, truth is stranger than fiction."}.
A match is found if: 1) the SkipThought Vector model's semantic relatedness score between the user's last utterance and a trigger phrase is above a predefined threshold, and 2) the user's last utterance contains keywords relevant to the trigger phrase.\footnote{Some trigger phrases do not have keywords. In this case, matching is based only on semantic relatedness.}
In total, there are $315$ trigger phrases (most are paraphrases of each other) and $35$ response sets.

If the model did not find a match in the first step, it proceeds to the second step.
In this step, the model selects its response from among all Reddit dataset responses.
As before, a set of $K$ model responses are retrieved using cosine similarity. The model then returns the response with the highest semantic relatedness score.

\textbf{Dual Encoder Models}:
The system contains two Dual Encoder retrieval models~\citep{lowe2015ubuntu,lowe2017training}, \emph{DualEncoderRedditPolitics} and \emph{DualEncoderRedditNews}.
Both models are composed of two sequence encoders $\textrm{ENC}_{Q}$ and $\textrm{ENC}_{R}$ with a single LSTM recurrent layer used to encode the dialogue history and a candidate response.
The score for a candidate response is computed by a bilinear mapping of the dialogue history embedding and the candidate response embedding as~\citet{lowe2015ubuntu}.
The models are trained using the method proposed by~\citep{lowe2015ubuntu}.
In principle, it is also possible to use early stopping based on separate model trained on a domain similar to our target domain~\citep{lowe2016evaluation}.
The response with the highest score from a set of $K=50$ candidate responses are retrieved using TF-IDF cosine similarity based on Glove word embeddings.
The model \emph{DualEncoderRedditPolitics} is trained on the Reddit \url{https://www.reddit.com/r/politics} dataset and extracts responses from all Reddit datasets.
The model \emph{DualEncoderRedditNews} is trained on the Reddit \url{https://www.reddit.com/r/news} dataset and extracts responses from all Reddit datasets.

\textbf{Bag-of-words Retrieval Models}:
The system contains three bag-of-words retrieval models based on TF-IDF Glove word embeddings~\citep{pennington2014glove} and Word2Vec embeddings~\citep{mikolov2013distributedbetter}.\footnote{We use the pre-trained Word2Vec embeddings: \url{https://code.google.com/archive/p/word2vec/}.}
Similar to the VHRED models, these models retrieve the response with the highest cosine similarity.
The \emph{BoWWashingtonPost} model retrieves user comments from WashingtonPost news articles using Glove word embeddings.
The model \emph{BoWTrump} retrieves responses from a set of Twitter tweets scraped from Donald Trump's profile: \url{https://twitter.com/realDonaldTrump}.
This model also uses Glove word embeddings and it only returns a response when at least one relevant keyword or phrase is found in the user's utterance (e.g.\@ when the word \textit{"Trump"} is mentioned by the user). The list of trigger keywords and phrases include: \textit{'donald'}, \textit{'trump'}, \textit{'potus'}, \textit{'president of the united states'}, \textit{'president of the us'}, \textit{'hillary'}, \textit{'clinton'}, \textit{'barack'}, and \textit{'obama'}.
The model \textit{BoWFactGenerator} retrieves responses from a set of about 2500 \textit{interesting} and \textit{fun} facts, including facts about animals, geography and history.
The model uses Word2Vec word embeddings.
The model \textit{BoWGameofThrones} retrieves responses from a set of quotes scraped from \url{https://twitter.com/ThroneQuotes} using Glove word embeddings. Tweets from this source were manually inspected and cleaned to remove any tweets that were not quotes from the series. As in the \emph{BoWTrump} model, we use a list of trigger phrases to determine if the model's output is relevant to the user's utterance. We populate this list with around 80 popular character names, place names and family names, which are large unique to the domain. We also added a few aliases to try and account for alternative speech transcriptions of these named entities. Some phrases include: \textit{'ned stark'}, \textit{'jon snow'}, \textit{'john snow'}, \textit{'samwell tarly'}, \textit{"hodor"}, \textit{"dothraki"} and so on.
\footnote{This model was implemented after the competition ended, but is included here for completeness.}

\subsection{Retrieval-based Logistic Regression}
\textbf{BoWEscapePlan}:
The system contains a response model, called \textit{BoWEscapePlan}, which returns a response from a set of 35 topic-independent, generic pre-defined responses, such as \textit{"Could you repeat that again"}, \textit{"I don't know"} and \textit{"Was that a question?"}.
Its main purpose is to maintain user engagement and keep the conversation going, when other models are unable to provide meaningful responses.
This model uses a logistic regression classifier to select its response based on a set of higher-level features.

To train the logistic regression classifier, we annotated ~$12,000$ user utterances and candidate response pairs for appropriateness on a Likert-type scale $1-5$.
The user utterances were extracted from interactions between Alexa users and a preliminary version of the system.
The candidate responses were sampled at random from \textit{BoWEscapePlan}'s response list.
The label collection and training procedure for the logistic regression model are similar to the procedures described in Section \ref{sec:model_selection_policy}.
The logistic regression model is trained with log-likelihood on a training set, with early-stopping on a development set, and evaluated on the testing set.
However, the trained model's performance was poor.
It obtained a Pearson correlation coefficient of $0.05$ and a Spearman’s rank correlation coefficient of $0.07$.
This indicates that the logistic regression model is only slightly better at selecting a topic-independent, generic response compared to selecting a response at uniform random.
Future work should investigate collecting more labeled data and pre-training the logistic regression model.

\subsection{Search Engine-based Neural Networks}
The system contains a deep classifier model, called \emph{LSTMClassifierMSMarco}, which chooses its response from a set of search engine results.
The system searches the web with the last user utterance as query, and retrieves the first 10 search snippets.
The retrieved snippets are preprocessed by stripping trailing words, removing unnecessary punctuation and truncating to the last full sentence. 
The model uses a bidirectional LSTM to separately map the last dialogue utterance and the snippet to their own embedding vectors.
The resulting two representations are concatenated and passed through an MLP to predict a scalar-value between $0-1$ indicating how appropriate the snippet is as a response to the utterance.

The model is trained as a binary classification model on the Microsoft Marco dataset with cross-entropy to predict the relevancy of a snippet given a user query~\citep{nguyen2016ms}.
Given a search query and a search snippet, the model must output one when the search snippet is relevant and otherwise zero.
Search queries and ground truth search snippets are taken as positive samples,
while other search snippets are selected at random as negative samples.
On this task, the model is able to reach a prediction accuracy of $72.96\%$ w.r.t.\@ the Microsoft Marco development set. 


The system is able to use search APIs from various search engines including Google, Bing, and AIFounded~\citep{aifounded}. In the current model, we choose Google as the search engine,
since qualitative inspection showed that this retrieved the most appropriate responses.

\subsection{Generation-based Neural Networks}
The system contains a generative recurrent neural network language model, called \textit{GRUQuestionGenerator}, which can generate follow-up questions word-by-word, conditioned on the dialogue history.
The input to the model consists of three components: a one-hot vector of the current word, a binary question label and a binary speaker label.
The model contains two GRU layers~\citep{cho2014learning} and softmax output layer.
The model is trained on Reddit Politics and Reddit News conversations, wherein posts were labeled as questions by detecting question marks.
We use the optimizer Adam~\citep{kingma2014adampublished}, and perform early stopping by checking the perplexity on the validation set
For generation, we first condition the model on a short question template (e.g.\@ \textit{"How about"}, \textit{“What about”}, \textit{“How do you think of”}, \textit{“What is your opinion of”}), and then generate the rest of the question by sampling from the model with the question label clamped to one.
The generation procedure stops once a question mark is detected.
Further, the length of the question is controlled by tuning the temperature of the softmax layer.
Due to speed requirements, only two candidate responses are generated and the best one w.r.t.\@ log-likelihood of the first 10 words is returned.

\section{Model Selection Policy} \label{sec:model_selection_policy}
After generating the candidate response set, the dialogue manager uses a \textit{model selection policy} to select the response it returns to the user.
The dialogue manager must select a response which increases the satisfaction of the user for the entire dialogue.
It must make a trade-off between immediate and long-term user satisfaction.
For example, suppose the user asks to talk about politics.
If the dialogue manager chooses to respond with a political joke, the user may be pleased for one turn.
Afterwards, however, the user may be disappointed with the system's inability to debate political topics.
Instead, if the dialogue manager chooses to respond with a short news story, the user may be less pleased for one turn.
However, the news story may influence the user to follow up with factual questions, which the system may be better adept at handling.
To make the trade-off between immediate and long-term user satisfaction, we consider selecting the appropriate response as a \textit{sequential decision making problem}.
This section describes five approaches to learn the model selection policy.
These approaches are all evaluated with real-world users in the next section.

We use the reinforcement learning framework~\citep{sutton1998reinforcement}.
The dialogue manager is an agent, which takes actions in an environment in order to maximize rewards.
For each time step $t=1,\dots,T$, the agent observes the dialogue history $h_t$ and must choose one of $K$ actions (responses): $a_t^1,\dots,a_t^K$.
After taking an action, the agent receives a reward $r_t$ and is transferred to the next state $h_{t+1}$ (which includes the user's next response).
Then, the agent is provided with a new set of $K$ actions: $a_{t+1}^1,\dots,a_{t+1}^K$.
The agent's goal is to maximize the discounted sum of rewards:
\begin{align}
R = \sum_{t=1}^T \gamma^t r_t,
\end{align}
which is referred to as the \textit{expected cumulative return} (or simply expected return).
The parameter $\gamma \in (0,1]$ is a discount factor.

An issue specific to our setting is that the set of actions changes depending on the state (dialogue history). This happens because the candidate responses are generated by response models, which also depend on the dialogue history.
In addition, the response models are not deterministic.
This means the set of candidate responses is likely to be different every time the agent encounters the same state $h_t$.\footnote{In general, since some response models only output responses for certain user utterances, the number of candidate responses also changes depending on the state.}
This is in contrast to certain reinforcement learning problems, such as learning to play Atari 2600 games, where the set of actions is fixed given the state.
To simplify notation, we will fix the number of actions to $K$ henceforth.


\textbf{Action-value Parametrization}: We use two different approaches to parametrize the agent's policy.
The first approach is based on an action-value function, defined by parameters $\theta$:
\begin{align}
Q_{\theta}(h_t, a_t^k) \in \mathbb{R} \quad \text{for} \ k=1,\dots,K, \label{eq:action_value_function}
\end{align}
which estimates expected return of taking action $a_t^k$ (candidate response $k$) given dialogue history $h_t$ and given that the agent will continue to use the same policy afterwards.
Given $Q_{\theta}$, the agent chooses the action with highest expected return:
\begin{align}
\pi_{\theta}(h_t) = \argmax_{k} \ Q_{\theta}(h_t, a_t^k). \label{eq:greedy_action_value_function}
\end{align}



The use of an action-value function for selecting dialogue responses is closely related to the recent work by \citet{lowe2017toward}, where a model is learned to predict the quality of a dialogue system response.
However, in our case, $Q_{\theta}$ is only conditioned on the dialogue context.
On the other hand, the model proposed by \citet{lowe2017toward} is conditioned both on the dialogue context and on a human reference response. 
The action-value function is also related to the the work by \citet{yu2016strategy}, who learn an evaluation model, which is used to train a reinforcement learning agent to select appropriate dialogue response strategies.

\textbf{Stochastic Policy Parametrization}: The second approach instead parameterizes the policy as a discrete distribution over actions.
Let $\theta$ be the parameters. 
The agent selects its action by sampling:
\begin{align}
\pi_{\theta}(a_t^k | h_t) = \dfrac{ e^{\lambda^{-1}f_{\theta}(h_t, a_t^k)}}{ \sum_{a_t'}  e^{\lambda^{-1}f_{\theta}(h_t, a_t')}} \quad \text{for} \ k=1,\dots,K, \label{eq:stochastic_policy}
\end{align}
where $f_{\theta}(h_t, a_t^k)$ is the \textit{scoring function}, which assigns a scalar score to each response $a_t^k$ given $h_t$.
The parameter $\lambda$ is called the temperature and controls the entropy of the distribution.
The higher $\lambda$ is, the more uniform the selecting of actions will be.
The stochastic policy can be transformed to a deterministic (greedy) policy by selecting the action with highest probability:
\begin{align}
\pi_{\theta}^{\text{greedy}}(h_t) = \argmax_{k} \ \pi_{\theta}(a_t^k | h_t) = \argmax_{k} \ f_{\theta}(h_t, a_t^k). \label{eq:greedy_stochastic_policy}
\end{align}


\textbf{Scoring Model}: The action-value function $Q_{\theta}(h_t, a_t^k)$ and scoring function $f_{\theta}(h_t, a_t^k)$ are closely related.
Both functions yield a ranking over the actions; higher values imply higher expected returns.
When $Q_{\theta}(h_t, a_t^k) = f_{\theta}(h_t, a_t^k)$, 
the action-value function policy in eq.\@ \eqref{eq:greedy_action_value_function} is equivalent to the greedy policy in eq.\@ \eqref{eq:greedy_stochastic_policy}.
For simplicity, we will use the same parametrization for both $Q_{\theta}(h_t, a_t^k)$ and $f_{\theta}(h_t, a_t^k)$.
Therefore, we let both functions take the same features as input and process them using the same neural network architecture.
We will refer to both functions as the \textit{scoring model}.

The next section describes the input features for the scoring model.

\subsection{Input Features}
As input to the scoring model we compute 1458 features based on the given dialogue history and candidate response.
The input features are based on a combination of word embeddings, dialogue acts, part-of-speech tags, unigram word overlap, bigram word overlap and model-specific features:
\begin{labeling}{Word embeddings of last user utterance:}
\item [Word embeddings of response:] Average of candidate response word embeddings~\citep{mikolov2013distributedbetter}.\footnote{We use the pre-trained Word2Vec embeddings: \url{https://code.google.com/archive/p/word2vec/}.}
\item [Word embeddings of last user utterance:] Average of the last user utterance word embeddings.
\item [Word embeddings of context:] Average of the word embeddings of the last six utterances in dialogue context.
\item [Word embedding of user context:] Average of the word embeddings of the last three user utterances in dialogue context.
\item [Word embedding similarity metrics:] The \emph{Embedding Average}, \emph{Embedding Extrema} and \emph{Embedding Greedy} similarity metrics described by \citet{liu2016not}. Each similarity metric is computed between 1) the last user utterance and candidate response, 2) the last six utterances in the dialogue and candidate response, 3) the last three user utterances in the dialogue and candidate response, 4) the last six utterances in the dialogue and candidate response with stop-words removed, and 5) the last three user utterances in the dialogue and candidate response with stop-words removed.
\item [Response model class:] A one-hot vector with size equal to the number of response models, where entry $i$ is equal to $1.0$ when candidate response was generated by the model class with index $i$.
\item [Part-of-speech response class:] The part-of-speech tags for candidate response is estimated using a maximum entropy tagger trained on the Penn Treebank corpus. The sequence of part-of-speech tags is then mapped to a one-hot vector, which constitutes the input feature.
\item [Dialogue act response model class:] The outer-product between a one-hot vector representing the dialogue act (we consider 10 types of dialogue acts) and a one-hot vector for indicating the model class~\citep{stolcke2000dialogue}.
\item [Word overlap:] $1.0$ when one or more non-stop-words overlap between candidate response and last user utterance, and otherwise zero.
\item [Bigram overlap short-term:] $1.0$ when a bigram (two consecutive tokens) exists both in the candidate response and in the last user utterance, and otherwise zero.
\item [Bigram overlap long-term:] $1.0$ when a bigram exists both in candidate response and in one of the last utterances in dialogue context, and otherwise zero.
\item [Named-entity overlap short-term:] $1.0$ when a named-entity (an upper-cased word, which is not a stop-word) exists both in candidate response and in the last user utterance, and otherwise zero.
\item [Named-entity overlap long-term:] $1.0$ when a named-entity exists both in candidate response and in one of the last utterances in dialogue context, and otherwise zero.
\item [Generic response:] $1.0$ when candidate response consists of only stop-words or words shorter than 3 characters, and otherwise zero.
\item [Wh-word response feature:] $1.0$ when candidate response contains a wh-word (e.g. \textit{what}, \textit{where}, and so on), and otherwise zero.
\item [Wh-word context:] $1.0$ when last user utterance contains a wh-word, and otherwise zero.
\item [Intensifier word response:] $1.0$ when candidate response contains an intensifier word (e.g. \textit{amazingly}, \textit{crazy}, and so on), and otherwise zero.
\item [Intensifier word context:] $1.0$ when last user utterance contains an intensifier word, and otherwise zero.
\item [Unigram response:] A set of binary features which are $1.0$ when candidate response contains a specific word (including the words \textit{I}, \textit{you} and \textit{thanks}), and otherwise zero.
\item [Negation response:] $1.0$ when candidate response contains a negation word, such as \textit{not} or \textit{n't}, and otherwise zero.
\item [Non-stop-words response:] $1.0$ when candidate response contains a non-stop-word, and otherwise zero.
\end{labeling}

We do not include features based on the confidences of the speech recognition system, for experimental reasons.
Speech recognition errors are a confounding factor in experiments with real-world users.
Speech recognition errors are likely to affect user satisfaction.
If features based on speech recognition confidences were included, one policy might learn to handle speech recognition errors better than another policy.
In turn, this could make that policy perform better w.r.t.\@ overall user satisfaction.
However, that would be an effect caused by the imperfect speech recognition system, and would not reflect user satisfaction under a perfect speech recognition system.
Excluding these features as input to the scoring model helps minimize this confounding effect.
Nevertheless, even if these features are excluded, it should be noted that speech recognition errors still constitute a substantial confounding factor in our later experiments.
Lastly, for the same reasons, none of the response models utilize speech recognition confidences.




In principle, it is possible to compute input features by encoding the dialogue context and candidate response using Recurrent Neural Networks (RNNs) or Convolutional Neural Networks (ConvNets)~\citep{socher2013recursive,blunsom2014convolutional,cho2014learning,yu2014deep,kiros2015skip}.
However, these models are known to require training on large corpora in order to achieve acceptable performance, which we do not have access to.
In addition, we need to keep the scoring model's execution time under 150ms.
Otherwise, the slowdown in the response time, could frustrate the user and lower the overall user satisfaction.
This rules out large RNNs and ConvNets for the Amazon Alexa Prize competition, since these would require more computational runtime. However, future dialogue systems utilizing larger datasets should consider large-scale models.

\subsection{Model Architecture}
\begin{figure}[ht]
  \centering
  \includegraphics[scale=0.27]{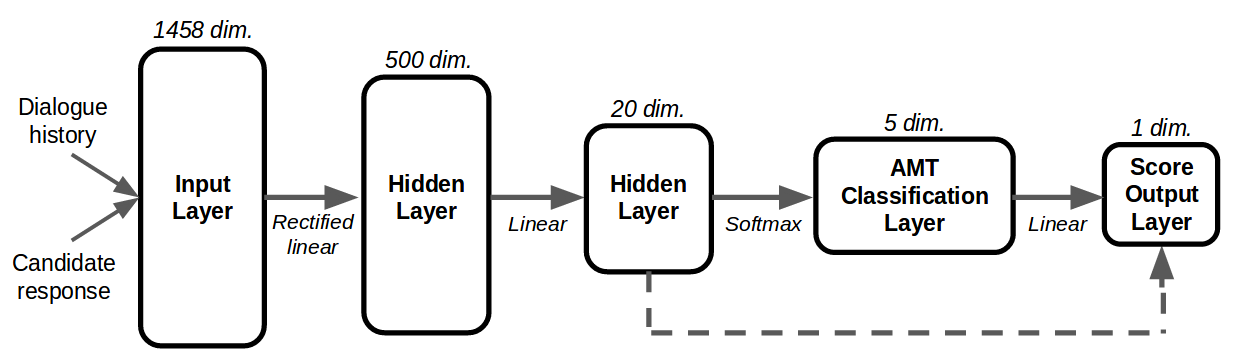}
  \caption{Computational graph for scoring model, used for the model selection policies based on both action-value function and stochastic policy parametrizations. The model consists of an input layer with $1458$ features, a hidden layer with $500$ hidden units, a hidden layer with $20$ hidden units, a softmax layer with $5$ output probabilities (corresponding to the five AMT labels in Section \ref{subsection:supervised_amt}), and a scalar-valued output layer. The dashed arrow indicates a skip connection.}
  \label{fig:scoring_model}
\end{figure}

This section describes the scoring model's architecture.
The scoring model is a five-layered neural network.
The first layer is the input, consisting of the 1458 features, described in the previous section.
The second layer contains 500 hidden units, computed by applying a linear transformation followed by the rectified linear activation function~\citep{nair2010rectified,glorot2011deep} to the input layer units.
The third layer contains 20 hidden units, computed by applying a linear transformation to the preceding layer units.
Similar to matrix factorization, this layer compresses the 500 hidden units down to 20 hidden units.
The fourth layer contains 5 outputs units, which are probabilities (i.e.\@ all values are positive and sum to one).
These output units are computed by applying a linear transformation to the preceding layer units followed by a softmax transformation.
This layer corresponds to the Amazon Mechanical Turk labels, which will be described in the next sub-section.
The fifth layer is the final output scalar, computed by applying a linear transformation to the units in the third and fourth layers.
The model is illustrated in Figure \ref{fig:scoring_model}.

Before settling on this architecture, we experimented both with deeper and more shallow models. However, we found that both the deeper models and the more shallow models performed worse.
Nevertheless, future work should explore alternative architectures.

We use five different machine learning approaches to learn the scoring model.
These are described next.

\subsection{Supervised AMT: Learning with Crowdsourced Labels} \label{subsection:supervised_amt}

This section describes the first approach to learning the scoring model,
which is based on estimating the action-value function using supervised learning on crowdsourced labels.
This approach also serves as initialization for the approaches discussed later.

\textbf{Crowdsourcing}: We use Amazon Mechanical Turk (AMT) to collect data for training the scoring model.
We follow a setup similar to \citet{liu2016not}.
We show human evaluators a dialogue along with 4 candidate responses, and ask them to score how appropriate each candidate response is on a 1-5 Likert-type scale.
The score 1 indicates that the response is inappropriate or does not make sense, 3 indicates that the response is acceptable, and 5 indicates that the response is excellent and highly appropriate. 

Our setup only asks human evaluators to rate the overall appropriateness of the candidate responses.
In principle, we could choose to evaluate other aspects of the candidate responses.
For example, we could evaluate fluency.
However, fluency ratings would not be very useful since most of our models retrieve their responses from existing corpora, which contain mainly fluent and grammatically correct responses.
As another example, we could evaluate topical relevancy.
However, we choose not to evaluate such criteria since it is known to be difficult to reach high inter-annotator agreement on them~\citep{liu2016not}.
In fact, it is well known that even asking for a single overall rating tends to produce only a fair agreement between human evaluators~\citep{charras2017comparing}; disagreement between annotators tends to arise either when the dialogue context is short and ambiguous, or when the candidate response is only partially relevant and acceptable.

The dialogues are extracted from interactions between Alexa users and preliminary versions of our system.
Only dialogues where the system does not have a priority response were extracted (when there is a priority response, the dialogue manager must always return the priority response).
About $3/4$ of these dialogues were sampled at random, and the remaining ~$1/4$ dialogues were sampled at random excluding identical dialogues.\footnote{Sampling at random is advantageous for our goal, because it ensures that candidate responses to frequent user statements and questions tend to be annotated by more turkers. This increases the average annotation accuracy for such utterances, which in turn increases the scoring model's accuracy for such utterances.}
For each dialogue, the corresponding candidate responses are created by generating candidate responses from the response models.

\begin{figure}[ht]
  \centering
  \fbox{\includegraphics[scale=0.22]{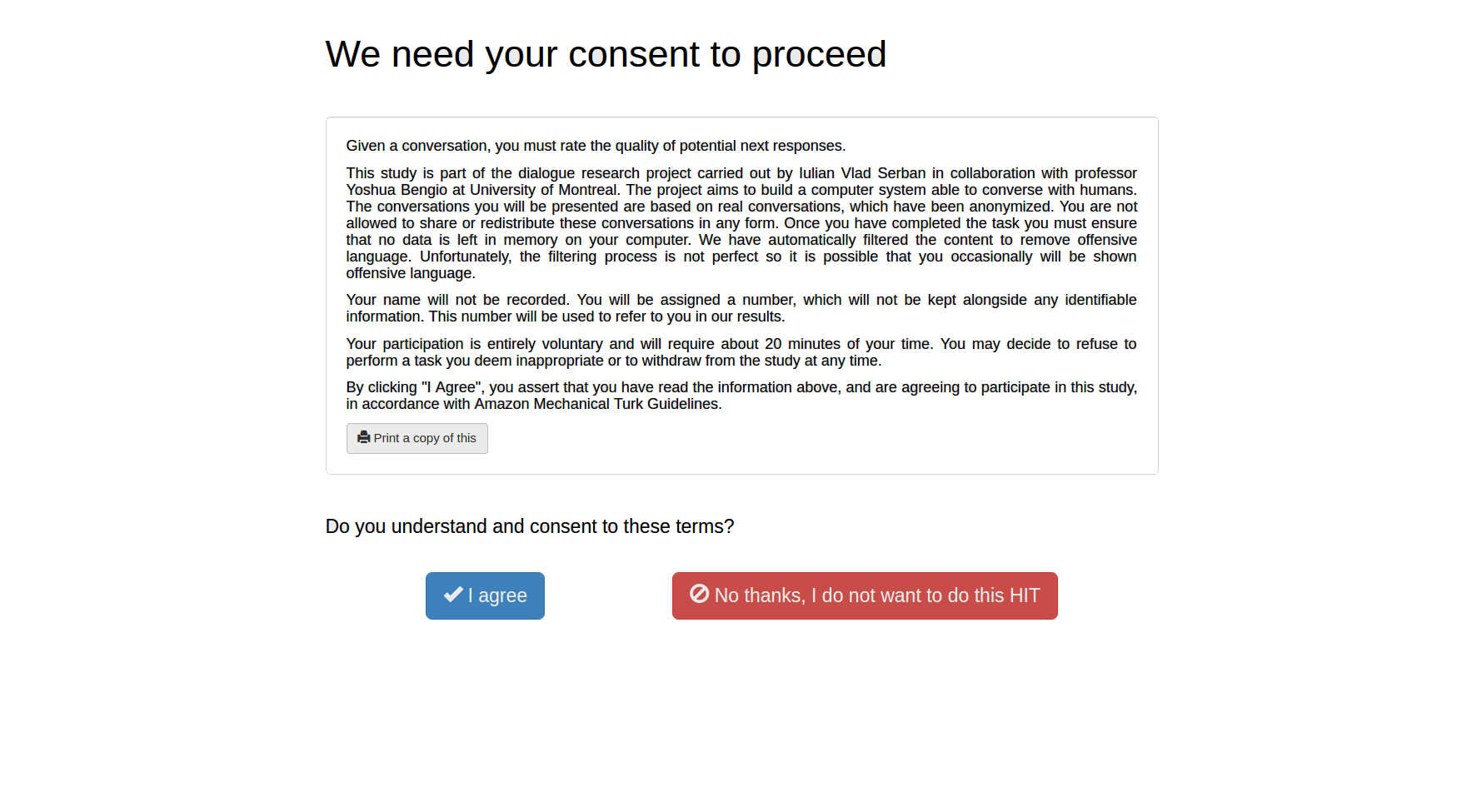}}
  \caption{Consent screen for Amazon Mechanical Turk human intelligence tasks (HITs).}
  \label{fig:amt_setup_consent}
\end{figure}

\begin{figure}[ht]
  \centering
  \fbox{\includegraphics[scale=0.22]{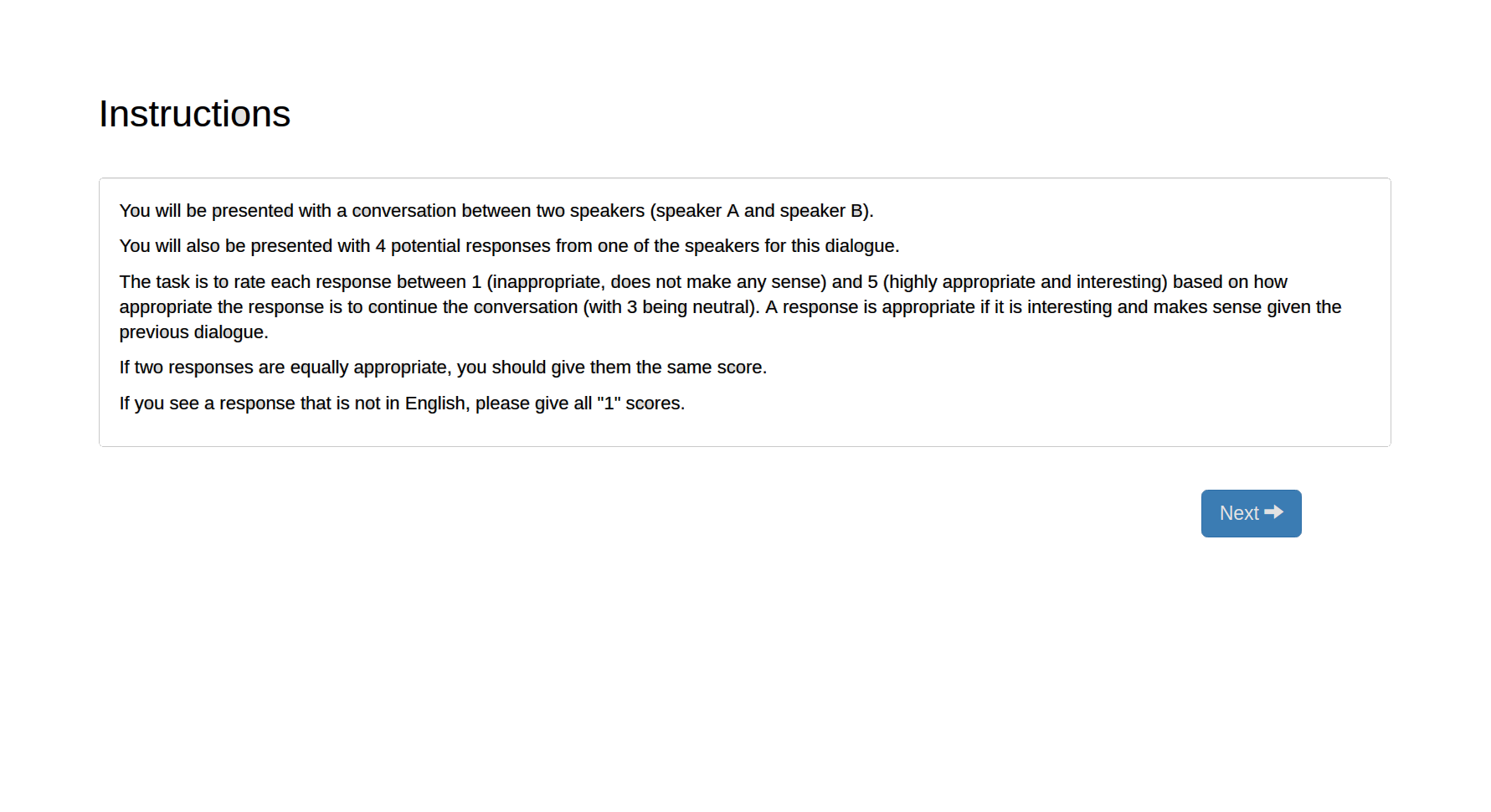}}
  \caption{Instructions screen for Amazon Mechanical Turk human intelligence tasks (HITs).}
  \label{fig:amt_setup_instructions}
\end{figure}

\begin{figure}[H]
  \centering
  \fbox{\includegraphics[scale=0.36]{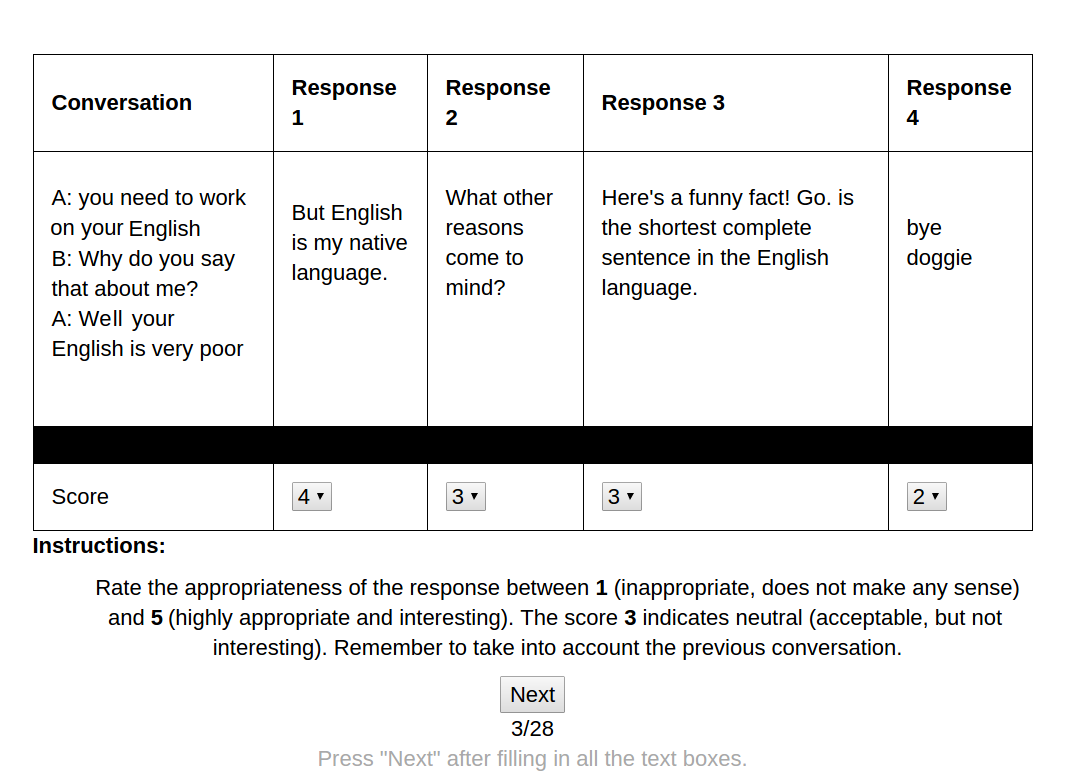}}
  \caption{Annotation screen for Amazon Mechanical Turk human intelligence tasks (HITs). The dialogue text is a fictitious example.}
  \label{fig:amt_setup_annotation}
\end{figure}

We preprocess the dialogues and candidate responses by masking out profanities and swear words with stars (e.g.\@ we map \textit{"fuck"} to \textit{"****"}).\footnote{The masking is not perfect.
Therefore, we also instruct turkers that the task may contain profane and obscene language. Further, it should also be noted that Amazon Mechanical Turk only employs adults.}
Furthermore, we anonymize the dialogues and candidate responses by replacing first names with randomly selected gender-neutral names (for example, \textit{"Hi John"} could be mapped to \textit{"Hello Casey"}).
Finally, the dialogues are truncated to the last 4 utterances and last 500 words.
This reduces the cognitive load of the annotators.
Examples from the crowdsourcing task are shown in Figure \ref{fig:amt_setup_consent}, Figure \ref{fig:amt_setup_instructions} and Figure \ref{fig:amt_setup_annotation}.
The dialogue example shown in Figure \ref{fig:amt_setup_annotation} is a fictitious example.

\newpage

We inspected the annotations manually.
We observed that annotators tended to frequently overrate topic-independent, generic responses.
Such responses may be considered acceptable for a single turn in a conversation, but are likely to be detrimental when repeated over and over again.
In particular, annotators tended to overrate responses generated by the response models \emph{Alicebot}, \emph{Elizabot}, \emph{VHREDSubtitles} and \emph{BoWEscapePlan}.
Responses generated by these models are often acceptable or good, but the majority of them are topic-independent, generic sentences.
Therefore, for these response models, we mapped all labels $5$ (\textit{"excellent"}) to $4$ (\textit{"good"}).
Furthermore, for responses consisting of only stop-words, we decreased the labels by one level (e.g.\@ $4$ is mapped to $3$).
Finally, the \emph{BoWMovies} response model suffered from a bug during the label collection period.
Therefore, we decreased all labels given to \emph{BoWMovies} responses to be at most $2$ (\textit{"poor"}).

In total, we collected $199,678$ labels.
We split this into training (train), development (dev) and testing (test) datasets consisting of respectively 137,549, 23,298 and 38,831 labels each.

\textbf{Training}: We optimize the scoring model w.r.t.\@ log-likelihood (cross-entropy) to predict the 4th layer, which represents the AMT label classes.
Formally, we optimize the parameters $\theta$:
\begin{align}
\hat{\theta} = \argmax_{\theta} \ \sum_{x,y} \log P_\theta(y | x), \label{eq:supervised_amt_model}
\end{align}
where $x$ are the input features, $y$ is the corresponding AMT label class (a one-hot vector) and $P_\theta(y | x)$ is the model's predicted probability of $y$ given $x$, computed in the second last layer of the scoring model.
We use the first-order gradient-descent optimizer Adam~\citep{kingma2014adampublished} 
We experiment with a variety of hyper-parameters, and select the best hyper-parameter combination based on the log-likelihood of the dev set.
For the first hidden layer, we experiment with layer sizes in the set: $\{500, 200, 50\}$.
For the second hidden layer, we experiment with layer sizes in the set: $\{50, 20, 5\}$.
We use L2 regularization on all model parameters, except for bias parameters.
We experiment with L2 regularization coefficients in the set: $\{10.0, 1.0, 10^{-1}, \dots, 10^{-9}\}$
Unfortunately, we do not have labels to train the last layer.
Therefore, we fix the parameters of the last layer to the vector $[1.0, 2.0, 3.0, 4.0, 5.0]$.
In other words, we assign a score of $1.0$ for the label \emph{very poor}, a score of $2.0$ for the label \emph{poor}, a score of $3.0$ for the label \emph{acceptable}, a score of $4.0$ for the label \emph{good} and a score of $5.0$ for the label \emph{excellent}.
As this model was trained on crowdsourced data from Amazon Mechanical Turk (AMT), we call this model \textit{Supervised AMT}.

\begin{table}[H]
  \caption{Scoring model evaluation on Amazon Mechanical Turk test set w.r.t.\@ Pearson correlation coefficient, Spearman's rank correlation coefficient and mean squared error.} \label{tabel:scoring_model_results}
  \small
  \centering
    \begin{tabular}{lccc}
    \toprule
     \textbf{Model} & \textbf{Pearson} & \textbf{Spearman} & \textbf{Mean squared error} \\
    \midrule
    \textit{Average Predictor} & $0.00$ & $0.00$ & $1.30$ \\
    \textit{Supervised AMT} & $\mathbf{0.40}$ & $\mathbf{0.38}$ & $\mathbf{1.10}$ \\ \bottomrule
    \end{tabular}
\end{table}

\begin{figure}[H]
  \centering
  \includegraphics[scale=0.375]{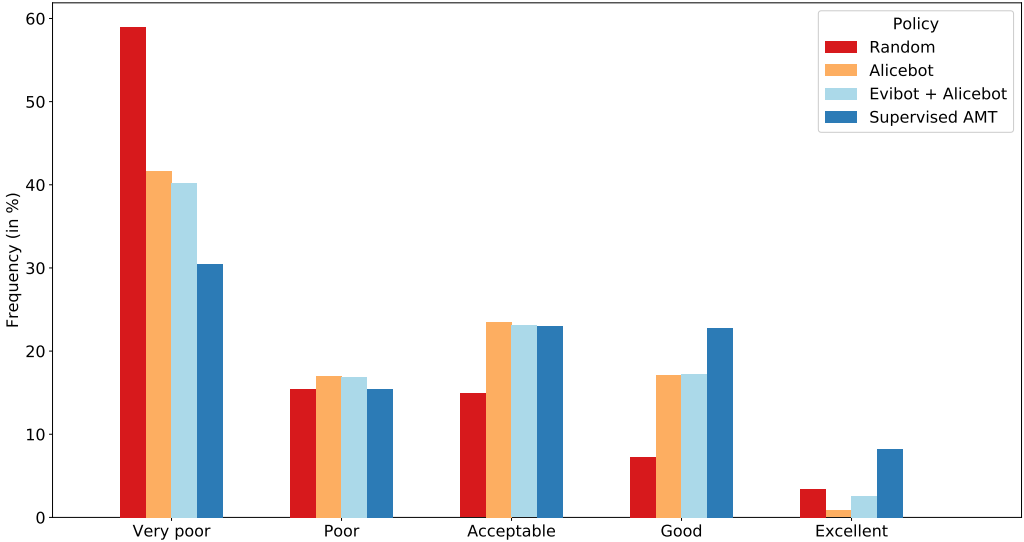}
  \caption{Amazon Mechanical Turk class frequencies on the test set w.r.t.\@ different policies.}
  \label{fig:amt_performance}
\end{figure}

Table \ref{tabel:scoring_model_results} shows the performance w.r.t.\@ Pearson correlation coefficient, Spearman's rank correlation coefficient and mean squared error.
The metrics are computed after linearly transforming the AMT class categories to the scalar output score (i.e.\@ by taking the dot-product between the one-hot class vector and the vector $[1.0, 2.0, 3.0, 4.0, 5.0]$).
The \emph{Average Predictor} is a baseline model, which always predicts with the average output score.
As shown, \emph{Supervised AMT} achieves a Pearson correlation coefficient of $0.40$, a Spearman's rank correlation coefficient of $0.38$ and a significant reduction in mean squared error.
This indicates \emph{Supervised AMT} performs significantly better than the baseline.

Figure \ref{fig:amt_performance} shows the performance w.r.t.\@ each AMT label class.
In addition to \emph{Supervised AMT}, the figure shows the performance of three baseline policies: 1) \emph{Random}, which selects a response at random, 2) \emph{Alicebot}, which selects an \emph{Alicebot} response if available and otherwise selects a response at random, and 3) \emph{Evibot + Alicebot}, which selects an \emph{Evibot} response if available and otherwise selects an \emph{Alicebot} response.
For each policy, the figure shows the percentage of responses selected by the policy belonging to a particular AMT label class.
In one end of the spectrum, we observe that \emph{Supervised AMT} has a \textasciitilde $30\%$ point reduction compared to \emph{Random} in responses belonging to the \textit{"very poor"} class.
For the same AMT label class, \emph{Supervised AMT} has a reduction of \textasciitilde $10\%$ points compared to \emph{Alicebot} and \emph{Evibot + Alicebot}.
In the other end of the spectrum, we observe that \emph{Supervised AMT} performs significantly better than the three baselines w.r.t.\@ the classes \textit{"good"} and \textit{"excellent"}.
In particular, \emph{Supervised AMT} reaches \textasciitilde $8\%$ responses belonging to the class \textit{"excellent"}.
This is more than double compared to all three baseline policies.
This demonstrates that \emph{Supervised AMT} has learned to select \textit{"good"} and \textit{"excellent"} responses, while avoiding \textit{"very poor"} and \textit{"poor"} responses.

Overall, the results show that \emph{Supervised AMT} improves substantially over all baseline policies.
Nevertheless, \textasciitilde $46\%$ of the \emph{Supervised AMT} responses belong to the classes \textit{"very poor"} and \textit{"poor"}.
This implies that there is ample space for improving both \emph{Supervised AMT} and the set of candidate responses (i.e.\@ the system's response models).


\subsection{Supervised Learned Reward: Learning with a Learned Reward Function}

In the first scoring model \emph{Supervised AMT} we fixed the last output layer weights to $[1.0, 2.0, 3.0, 4.0, 5.0]$.
In other words, we assigned a score of $1.0$ for \emph{very poor} responses, $2.0$ for \emph{poor} responses, $3.0$ for \emph{acceptable} responses, and so on.
It's not clear whether this score is correlated with scores given by real-world Alexa users, which is what we ultimately want to optimize the system for.
This section describes another approach, which remedies this problem by learning to predict the Alexa user scores based on previously recorded dialogues.

\textbf{Learned Reward Function}:
Let $h_t$ be a dialogue history and let $a_t$ be the corresponding response, given by the system at time $t$.
We aim to learn a linear regression model, $g_{\phi}$, which predicts the corresponding return (Alexa user score) at the current dialogue turn:
\begin{align}
g_{\phi}(h_t, a_t) \in [1, 5],
\end{align}
where $\phi$ are the model parameters.
We call this a \textit{reward model}, since it directly models the Alexa user score, which we aim to maximize.

Let $\{h_t^d, a_t^d, R^d\}_{d, t}$ be a set of examples, where $t$ denotes the time step and $d$ denotes the dialogue.
Let $R^d \in [1, 5]$ denote the observed real-valued return for dialogue $d$.

Specifically, we set $R^d$ to be the Alexa user score given at the end of dialogue $d$.
It's optional for users to a give a score; users are prompted to give a score at the end, but they may opt out by stopping the application.
Although not all users give scores, we do not consider examples without scores.\footnote{By ignoring dialogues without Alexa user scores, we introduce a significant bias in our reward model. In particular, it seems likely that the users who did not provide a score either found the system to be very poor or to lack particular functions/features they expected (e.g.\@ \textit{non-conversational activities}, such as playing games or taking quizzes.). A related problem arises in medical statistics, when patients undergo a treatment and, later, their outcome is not observed.}
Furthermore, users are encouraged to give a score in the range $1-5$.
The majority of users give whole number (integer) scores, but some users give decimal scores (e.g.\@ $3.5$).
Therefore, we treat $R^d$ as a real-valued number in the range $1-5$.

We learn $\phi$ by minimizing the squared error between the model's prediction and the observed return:
\begin{align}
\hat{\phi} = \argmax_{\phi} \ \sum_{d} \sum_{t} (g_{\phi}(h_t^d, a_t^d) - R^d)^2
\end{align}
As before, we optimize the model parameters with mini-batch stochastic gradient descent (SGD) using Adam.
We use L2 regularization with coefficients in the set $\{10.0, 1.0, 0.1, 0.01, 0.001, 0.0001, 0.00001, 0.0\}$. We select the coefficient with the smallest squared error on a hold-out dataset.

As input to the reward model we compute 23 features based on the dialogue history and a candidate response.
As training data is scarce, we use only higher-level features:
\begin{labeling}{Generic user utterance:}
\item [AMT label class:] A vector indicating the probability of the AMT label classes for the candidate response, computed using \emph{Supervised AMT}, as well as the probability that the candidate response has priority. If the candidate response has priority, the vector is zero in all entries, except the last entry corresponding to the priority class: $[0.0, 0.0, 0.0, 0.0, 0.0, 1.0]$.
\item [Generic response:] A binary feature, which is $1.0$ when the response only contains stop-words and otherwise zero.
\item [Response length:] The number of words in the response, and the square root of the number of words in the response.
\item [Dialogue act:] A one-hot vector, indicating whether the last user utterance's dialogue is a \textit{request}, a \textit{question}, a \textit{statement} or contains profanity~\citep{stolcke2000dialogue}.
\item [Sentiment class:] A one-hot vector, indicating whether the last user utterance's dialogue is negative, neutral or positive.
\item [Generic user utterance:] A binary feature, which is $1.0$ when the last user utterance only contains stop-words, and otherwise zero.
\item [User utterance length:] The number of words in the last user utterance, and the square root of the number of words in the response.
\item [Confusion indicator:] A binary feature, which is $1.0$ when the last user utterance is very short (less than three words) and contains at least one word indicating the user is confused (e.g.\@ \textit{"what"}, \textit{"silly"}, \textit{"stupid"}).
\item [Dialogue length:] The number of dialogue turns so far, as well as the square root and logarithm of the number of dialogue turns.
\end{labeling}

In total, our dataset for training the reward model has 4340 dialogues.
We split this into a training set with 3255 examples and a test set with 1085 examples.

To increase data efficiency, we learn an ensemble model through a variant of the bagging technique~\citep{breiman1996bagging}.
We create 5 new training sets, which are shuffled versions of the original training set.
Each shuffled dataset is split into a sub-training set and sub-hold-out set.
The sub-hold-out sets are created such that the examples in one set do not overlap with other sub-hold-out sets.
A reward model is trained on each sub-training set, with its hyper-parameters selected on the sub-hold-out set.
This increases data efficiency by allowing us to re-use the sub-hold-out sets for training, which would otherwise not have been used.
The final reward model is an ensemble, where the output is an average of the underlying linear regression models.

The reward model obtains a mean squared error of $0.96$ and a Spearman's rank correlation coefficient of $0.19$ w.r.t.\@ the real Alexa user on the test set.
In comparison, a model predicting with the average user score obtains a mean squared error of $0.99$ and (because it outputs a constant) a Spearman's rank correlation coefficient of zero.
Although the reward model is better than predicting the average, its correlation is relatively low.
There are two reasons for this.
First, the amount of training data is very small.
This makes it difficult to learn the relationships between the features and the Alexa user scores.
Second, the Alexa user scores are likely to have high variance because, they are influenced by many different factors.
The score of the user may be determined by a single turn in the dialogue (e.g.\@ a single misunderstanding at the end of the dialogue could result in a very low user score, even if all the previous turns in the dialogue were excellent).
The score of the user may be affected by the accuracy of the speech recognition module.
More speech recognition errors will inevitably lead to frustrated users.
In a preliminary study, we found that Spearman's rank correlation coefficient between the speech recognition confidences and the Alexa user scores was between $0.05 - 0.09$.
In comparison to correlations with other factors, this implies that speech recognition performance plays an important role in determining user satisfaction.\footnote{This was confirmed by manual inspection of the conversation logs, where the majority of conversations had several speech recognition errors. In conversations with an excessive number of speech recognition errors (perhaps due to noisy environments), the users' utterances clearly showed frustration with the system.}
In addition, extrinsic factors are likely to have a substantial influence on the user scores.
The user scores are likely to depend not only on the dialogue, but also on the user's profile (e.g.\@ whether the user is an adult or a child), the environment (e.g.\@ whether the user is alone with the system or several users are taking turns conversing with the system), the user's expectations towards the system before starting the conversation (e.g.\@ whether the system is capable of playing games) and the emotional state of the user (e.g.\@ the user's mood).

\textbf{Training}:
To prevent overfitting, we do not train the scoring model (action-value function) from scratch with the reward model as target.
Instead, we first initialize the model with the parameters of the \textit{Supervised AMT} scoring model, and then fine-tune it with the reward model outputs to minimize the squared error:
\begin{align}
\hat{\theta} = \argmax_{\theta} \sum_{d} \sum_{t} (f_{\theta}(h_t^d, a_t^d) - g_{\phi}(h_t^d, a_t^d))^2,
\end{align}
As before, we optimize the model parameters with stochastic gradient descent using Adam.
As training this model does not depend on AMT labels, training is carried out on recorded dialogues.
We train on several thousand recorded dialogue examples, where about $80\%$ are used for training and about $20\%$ are used as hold-out set.
No regularization is used.
We early stop on the squared error of the hold-out dataset  w.r.t.\@ Alexa user scores predicted by the reward model.
As this scoring model was trained with a learned reward function, we call it \textit{Supervised Learned Reward}.

\subsection{Off-policy REINFORCE}
As discussed earlier, one way to parametrize the policy is as a discrete probability distribution over actions.
This parametrization allows us to learn the policy directly from recorded dialogues through a set of methods known as \textit{policy gradient} methods.
This section describes one such approach.

\textbf{Off-policy Reinforcement Learning}: We use a variant of the classical \textit{REINFORCE} algorithm~\citep{williams1992simple,precup2000eligibility,precup2001off}, which we call \emph{Off-policy REINFORCE}.
Recall eq.\@ \eqref{eq:stochastic_policy}, where the policy's distribution over actions is parametrized as softmax function applied to a function $f_{\theta}$ with parameters $\theta$.
As before, let $\{h_t^d, a_t^d, R^d\}_{d, t}$ be a set of examples, where $h_t^d$ is the dialogue history for dialogue $d$ at time $t$, $a_t^d$ is the agent's action for dialogue $d$ at time $t$ and $R^d$ is the return for dialogue $d$.
Let $D$ be the number of dialogues and let $T^d$ be the number of turns in dialogue $d$.
Further, let $\theta_d$ be the parameters of the stochastic policy $\pi_{\theta_d}$ used during dialogue $d$.
The \emph{Off-policy REINFORCE} algorithm updates the policy parameters $\theta$ by:
\begin{align}
\Delta \theta \ \propto \ c_{t}^d \ \nabla_{\theta} \log \pi_{\theta}(a_t^d | h_t^d) \ R^d \quad \text{where} \ d \sim \text{Uniform}(1, D) \ \text{and} \ t \sim \text{Uniform}(1, T^d), \label{eq:offpolicy_reinforce}
\end{align}
where $c_{t}^d$ is the importance weight ratio:
\begin{align}
c_{t}^d \defeq \dfrac{\prod_{t'=1}^t \pi_{\theta}(a_{t'}^d | h_{t'}^d)}{\prod_{t'=1}^t \pi_{\theta_d}(a_{t'}^d | h_{t'}^d)}.
\end{align}
This ratio corrects for the discrepancy between the learned policy $\pi_{\theta}$ and the policy under which the data was collected $\pi_{\theta_d}$ (sometimes referred to as the behaviour policy).
It up-weights examples with high probability under the learned policy and down-weights examples with low probability under the learned reward function.

The intuition behind the algorithm can be illustrated by analogy with learning from trial and error. 
When an example has a high return (i.e.\@ high user score), the term $\nabla_{\theta} \log \pi_{\theta}(a_t^d | h_t^d) \ R^d$ will be a vector pointing in a direction increasing the probability of taking action $a_t^d$.
On the other hand, when an example has low return (i.e.\@ low user score), the term $\nabla_{\theta} \log \pi_{\theta}(a_t^d | h_t^d) \ R^d$ will be a vector close to zero or a vector pointing in the opposite direction, hence decreasing the probability of taking action $a_t^d$.

The importance ratio $c_{t}^d$ is known to exhibit very high, possibly infinite, variance~\citep{precup2001off}.
Therefore, we truncate the products in the nominator and denominator to only include the current time step $t$:
\begin{align}
c_{t,\text{trunc.\@}}^d \defeq \dfrac{\pi_{\theta}(a_{t}^d | h_{t}^d)}{\pi_{\theta_d}(a_{t}^d | h_{t}^d)}.
\end{align}
This induces bias in the learning process, but also acts as a regularizer.

\textbf{Reward Shaping}:
As mentioned before, one problem with the \emph{Off-policy REINFORCE} algorithm presented in eq.\@ \eqref{eq:offpolicy_reinforce} is that it suffers from high variance~\citep{precup2001off}.
The algorithm uses the return, observed only at the very end of an episode, to update the policy's action probabilities for all intermediate actions in an episode.
With a small number of examples, the variance in the gradient estimator is overwhelming and this could easily lead the agent to over-estimate the utility of poor actions and, vice versa, to under-estimate the utility of good actions.
One remedy for this problem is \textit{reward shaping}, where the reward at each time step is estimated using an auxiliary function~\citep{ng1999policy}.
For our purpose, we propose a simple variant of reward shaping which takes into account the sentiment of the user.
When the user responds with a negative sentiment (e.g.\@ an angry comment), we will assume that the preceding action was highly inappropriate and assign it a reward of zero.
Given a dialogue $d$, at each time $t$ we assign reward $r_t^d$:
\begin{align}
r_t^d \defeq \begin{cases} 
      0 & \text{if user utterance at time $t+1$ has negative sentiment,}  \\
      \dfrac{R^d}{T^d} & \text{otherwise.}
       \end{cases}
\end{align}
With reward shaping and truncated importance weights, the learning update becomes:
\begin{align}
\Delta \theta \propto c_{t,\text{trunc.\@}}^d \nabla_{\theta} \log \pi_{\theta}(a_t^d | h_t^d) \ r_t^d \quad \text{where} \ d \sim \text{Uniform}(1, D), t \sim \text{Uniform}(1, T^d), \label{eq:offpolicy_reinforce_reward_shaping}
\end{align}

\textbf{Off-policy Evaluation}:
To evaluate the policy, we estimate the expected return~\citep{precup2000eligibility}:
\begin{align}
 \text{R}_{\pi_{\theta}}[R] \ \approx \ \sum_{d,t} c_{t,\text{trunc.\@}}^d \ r_t^d. \label{eq:offpolicy_reinforce_evaluation}
\end{align}
Furthermore, by substituting $r_t^d$ with a constant reward of $1.0$ for each time step, we can compute the estimated number of time steps per episode under the policy.
As will be discussed later, this is an orthogonal metric based on which we can analyse and evaluate each policy.
However, this estimate does not include the number of priority responses, since there are no actions for the agent to take when there is a priority response.


\textbf{Training}:
We initialize the policy model with the parameters of \textit{Supervised AMT}, and then train the parameters w.r.t.\@ eq.\@ \eqref{eq:offpolicy_reinforce_reward_shaping} with stochastic gradient descent using Adam.
We use a set of a few thousand dialogues recorded between Alexa users and a preliminary version of the system.
About $60\%$ of these examples are used for training, and about $20\%$ are used for development and testing.
To reduce the risk of overfitting, we only train the weights related to the second last layer using \textit{off-policy REINFORCE}.
We use a random grid search with different hyper-parameters, which include the temperature parameter $\lambda$ and the learning rate.
We select the hyper-parameters with the highest expected return on the development set.


\subsection{Off-policy REINFORCE with Learned Reward Function}
Similar to the \emph{Supervised Learned Reward} policy, we may use the reward model for training with the \emph{Off-policy REINFORCE} algorithm.
This section describes how we combine the two approaches.

\textbf{Reward Shaping with Learned Reward Model}:
We use the reward model to compute a new estimate for the reward at each time step in each dialogue:
\begin{align}
r_t^d \defeq \begin{cases} 
      0 & \text{if user utterance at time $t+1$ has negative sentiment,}  \\
      g_{\phi}(h_t, a_t) & \text{otherwise.}
       \end{cases}
\end{align}
This is substituted into eq.\@ \eqref{eq:offpolicy_reinforce_reward_shaping} for training and into eq.\@ \eqref{eq:offpolicy_reinforce_evaluation} for evaluation.

\textbf{Training}:
As with \emph{Off-policy REINFORCE}, we initialize the policy model with the parameters of the \textit{Supervised AMT} model, and then train the parameters w.r.t.\@ eq.\@ \eqref{eq:offpolicy_reinforce_reward_shaping} with mini-batch stochastic gradient descent using Adam.
We use the same set of dialogues and split as \emph{Off-policy REINFORCE}.
We use a random grid search with different hyper-parameters,
As before, to reduce the risk of overfitting, we only train the weights related to the second last layer using this method.
which include the temperature parameter $\lambda$ and the learning rate, and select the hyper-parameters with the highest expected return on the development set.
In this case, the expected return is computed according to the learned reward model.
As this policy uses the learned reward model, we call it \emph{Off-policy REINFORCE Learned Reward}.

\subsection{Q-learning with the Abstract Discourse Markov Decision Process}
The approaches described so far have each their own advantages and disadvantages.
One way to quantify their differences is through a decomposition known as the \textit{bias-variance trade-off}.
At one end of the spectrum, the \emph{Supervised AMT} policy has low variance, because it was trained with hundreds of thousands of human annotations at the level of each model response.
However, for the same reason, \emph{Supervised AMT} incurs a substantial bias, because the human annotations do not reflect the real user satisfaction for an entire conversation.
At the other end of the spectrum, \emph{Off-policy REINFORCE} suffers from high variance, because it was trained with only a few thousand dialogues and corresponding user scores.
To make matters worse, the user scores are affected by many external factors (e.g.\@ user profile, user expectations, and so on) and occur at the granularity of an entire conversation.
Nevertheless, this method incurs low bias because it directly optimizes the objective metric we care about (i.e.\@ the user score).\footnote{Due to truncated importance weights, however, the \textit{off-policy REINFORCE} training procedure is still biased.}
By utilizing a learned reward function, \emph{Supervised Learned Reward} and \emph{Off-policy REINFORCE Learned Reward} suffer less from bias, but since the learned reward function has its own variance component, they are both bound to have higher variance.
In general, finding the optimal trade-off between bias and variance can be notoriously difficult.
In this section we propose a novel method for trading off bias and variance by learning the policy from simulations in an approximate Markov decision process.

\textbf{Motivation}
A Markov decision process (MDP) is a framework for modeling sequential decision making~\citep{sutton1998reinforcement}.
In the general setting, an MDP is a model consisting of a discrete set of states $H$, a discrete set of actions $A$, a transition distribution function $P$, a reward distribution function $R$, and a discount factor $\gamma$.
As before, an agent aims to maximize its reward during each episode.
Let $t$ denote the time step of an episode with length $T$.
At time step $t$, the agent is in state $h_t \in H$ and takes action $a_t \in A$.
Afterwards, the agent receives reward $r_t \sim R(h_t, a_t)$ and transitions to a new state $h_{t+1} \sim P(h_t | a_t)$.

Given an MDP model for open-domain conversations, there are dozens of algorithms we could apply to learn the agent's policy~\citep{sutton1998reinforcement}.
Unfortunately, such an MDP is difficult to build or estimate.
We could try to naively estimate one from the recorded dialogues, but this would require solving two extremely difficult problems.
First, we would need to learn the transition distribution $P$, which outputs the next user utterance in the dialogue given the dialogue history.
This problem is likely to be as difficult as our original problem of finding an appropriate response to the user!
Second, we would need to learn the reward distribution $R$ for each time step.
However, as we have shown earlier, it is very difficult to learn to predict the user score for an entire dialogue.
Given the data we have available, estimating the reward for a single turn is likely also going to be difficult.
Instead, we propose to tackle the problem by splitting it into three smaller parts.

\begin{figure}[ht]
  \centering
  \includegraphics[scale=0.20]{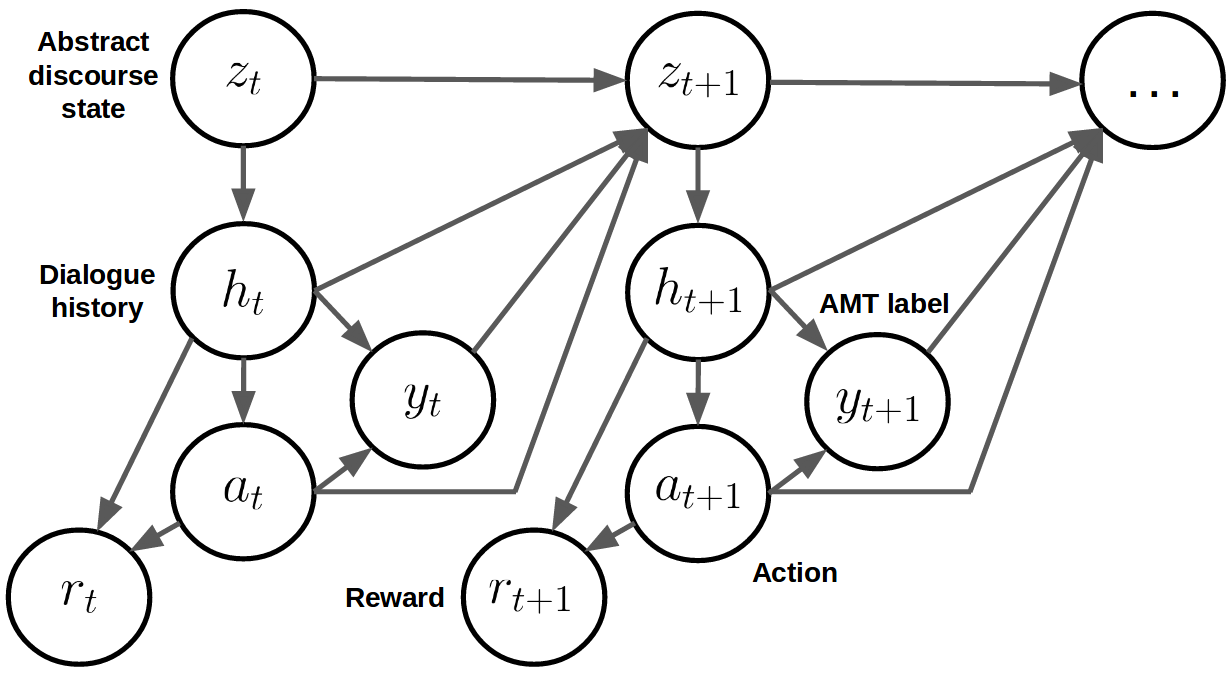}
  \caption{Probabilistic directed graphical model for the \emph{Abstract Discourse Markov Decision Process}. For each time step $t$, $z_t$ is a discrete random variable which represents the abstract state of the dialogue, $h_t$ represents the dialogue history, $a_t$ represents the action taken by the system (i.e.\@ the selected response), $y_t$ represents the sampled AMT label and $r_t$ represents the sampled reward.}
  \label{fig:approximate_mdp}
\end{figure}

\textbf{The Abstract Discourse Markov Decision Process}
The model we propose to learn is called the \emph{Abstract Discourse MDP}.
As illustrated in Figure \ref{fig:approximate_mdp}, the model follows a hierarchical structure at each time step.
At time $t$, the agent is in state $z_t \in Z$, a discrete random variable representing the \textit{abstract discourse state}.
This variable only represents a few high-level properties related to the dialogue history.
We define the set $Z$ is the Cartesian product:
\begin{align}
Z = Z_\text{Dialogue act} \times Z_\text{User sentiment} \times Z_\text{Generic user utterance},
\end{align}
where $Z_\text{Dialogue act}$, $Z_\text{User sentiment}$ and $Z_\text{Generic user utterance}$ are three discrete sets.
The first set consists of $10$ dialogue acts: $Z_\text{Dialogue act} = \{\text{Accept}, \text{Reject}, \text{Request}, \text{Politics}, \text{Generic Question}, \text{Personal Question}, \text{Statement}, \text{Greeting}, \\  \text{Goodbye}, \text{Other}\}$.
These dialogue acts represent the high-level intention of the user's utterance~\citep{stolcke2000dialogue}.
The second set consists of sentiments types: $Z_\text{User sentiment} = \{\text{Negative}, \text{Neutral}, \text{Positive}\}$.
The third set represent a binary variable: $Z_\text{Generic user utterance} = \{\text{True}, \text{False}\}$.
This variable is \textit{True} only when the user utterance is generic and topic-independent (i.e.\@ when the user utterance only contains stop-words).
We build a hand-crafted deterministic classifier, which maps a dialogue history to the corresponding classes in $Z_\text{Dialogue act}$, $Z_\text{User sentiment}$ and $Z_\text{Generic user utterance}$.
We denote this mapping $f_{h \to z}$.
Although we only consider dialogue acts, sentiment and generic utterances, it is trivial to expand the \textit{abstract discourse state} with other types of discrete or real-valued variables.

Given a sample $z_t$, the \emph{Abstract Discourse MDP} samples a dialogue history $h_t$ from a finite set of dialogue histories $H$.
In particular, $h_t$ is sampled at uniformly random from the set of dialogue histories where the last utterance is mapped to $z_t$:
\begin{align}
h_t \sim P(h | H, f_{h \to z}, z_t) \defeq \text{Uniform}(\{h \ | \ h \in H \ \text{and} \ f_{h \to z}(h) = z_t \}).
\end{align}
In other words, $h_t$ is a dialogue history where dialogue act, user sentiment and generic property is identical to the discrete variable $z_t$.

For our purpose, $H$ is the set of all recorded dialogues between Alexa users and a preliminary version of the system.
This formally makes the \emph{Abstract Discourse MDP} a \textit{non-parametric} model, since sampling from the model requires access to the set of recorded dialogue histories $H$.
This set grows over time when the system is deployed in practice.
This is useful, because it allows to continuously improve the policy as new data becomes available.
Further, it should be noted that the set $Z$ is small enough that every possible state is observed several times in the recorded dialogues.

Given a sample $h_t$, the agent chooses an action $a_t$ according to its policy $\pi_{\theta}(a_t | h_t)$, with parameters $\theta$.
A reward $r_t$ is then sampled such that $r_t \sim R(h_t, a_t)$, where $R$ is a distribution function.
In our case, we use the probability function $P_{\hat{\theta}}$, where the parameters $\hat{\theta}$  are estimated using supervised learning on AMT labels in eq.\@ \eqref{eq:supervised_amt_model}.
We specify a reward of $-2.0$ for a \textit{"very poor"} response class, a reward of $-1.0$ for a \textit{"poor"} response class, a reward of $0.0$ for an \textit{"acceptable"} response class, a reward of $1.0$ for a \textit{"good"} response class and a reward of $2.0$ for an \textit{"excellent"} response class.
To reduce the number of hyperparameters, we use the expected reward instead of a sample:\footnote{For example, if we were to use a Gaussian distribution, we would have to at least also specify the variance parameter.}
\begin{align}
r_t = P_{\hat{\theta}}(y | h_t, a_t)^{\text{T}} [-2.0, -1.0, 0.0, 1.0, 2.0].
\end{align}

Next, a variable $y_t \in \{\text{"very poor"}, \text{"poor"}, \text{"acceptable"}, \text{"good"}, \text{"excellent"}\}$ is sampled:
\begin{align}
y_t \sim P_{\hat{\theta}}(y | h_t, a_t).
\end{align}
This variable represents one \textit{appropriateness interpretation} of the output.
This variable helps predict the future state $z_{t+1}$, because the overall appropriateness of a response has a significant impact on the user's next utterance (e.g.\@ very poor responses often cause users to respond with \textit{What?} or \textit{I don't understand.}).

Finally, a new state $z_{t+1}$ is sampled according to $P_{\hat{\psi}}$:
\begin{align}
z_{t+1} \sim P_{\hat{\psi}}(z | z_t, h_t, a_t, y_t).
\end{align}
where $P_{\hat{\psi}}$ is the transition distribution with parameters $\hat{\psi}$.
The transition distribution is parametrized by three independent two-layer MLP models, which take as input the same features as the scoring function, as well as 1) a one-hot vector representing the sampled response class $y_t$, 2) a one-hot vector representing the dialogue act of the last user utterance, 3) a one-hot vector representing the sentiment of the last user utterance, 4) a binary variable indicating whether the last user utterance was generic, and 5) a binary variable indicating whether the last user utterance contained a wh-word (e.g.\@ \textit{what}, \textit{who}).
The first MLP predicts the next dialogue act, the second MLP predicts the next sentiment type and the third MLP predicts whether the next user utterance is generic.
The dataset for training the MLPs consists of $499,757$ transitions, of which $70\%$ are used for training and $30\%$ for evaluation.
The MLPs are trained with maximum log-likelihood using mini-batch stochastic gradient descent.
We use Adam and early-stop on a hold-out set.
Due to the large number of examples, no regularization is used.
The three MLP models obtain a joint perplexity of $19.51$.
In comparison, a baseline model, which always assigns the average class frequency as the output probability obtains a perplexity of $23.87$.
On average, this means that roughly $3-4$ possible $z_{t+1}$ states can be eliminated by conditioning on the previous variables $z_t, h_t, a_t$ and $y_t$.
In other words, the previous state $z_t$ and $h_t$, together with the agent's action $a_t$ has a significant effect on the future state $z_{t+1}$.
This means that an agent trained in the \emph{Abstract Discourse MDP} has the potential to learn to take into account future states of the dialogue when selecting its action.
This is in contrast to policies learned using supervised learning, which do not consider future dialogue states.

The idea of modeling a high-level abstraction of the dialogue, $z_t$, is related to the dialogue state tracking challenge~\citep{williams2013dialog,williams2016introduction}.
In this challenge, the task is to map the dialogue history to a discrete state representing all salient information about the dialogue.
Unlike the dialogue state tracking challenge, however, the variable $z_t$ only includes limited salient information about the dialogue.
For example, in our implementation, $z_t$ does not include topical information.
As such, $z_t$ is only a partial representation of the dialogue history.

\textbf{Training}
Given the \emph{Abstract Discourse MDP}, we are now able to learn policies directly from simulations.
We use \textit{Q-learning} with \textit{experience replay} to learn the policy parametrized as an action-value function~\citep{mnih2013playing,lin1993reinforcement}.
Q-learning is a simple off-policy reinforcement learning algorithm, which has been shown to be effective for training policies parametrized by neural networks.
For experience replay, we use a memory buffer of size $1000$.
We use an $\epsilon$-greedy exploration scheme with $\epsilon=0.1$.
We experiment with discount factors $\gamma \in \{0.1, 0.2, 0.5\}$.
As before, the parameters are updated using Adam.
To reduce the risk of overfitting, we only train the weights related to the final output layer and the skip-connection (shown in dotted lines in Figure \ref{fig:scoring_model}) using Q-learning.

Training is carried out in two alternating phases.
We train the policy for $100$ episodes.
Then, we evaluate the policy for $100$ episodes w.r.t.\@ average return.
Afterwards, we continue training the policy for another $100$ episodes.
During evaluation, each dialogue history is sampled from a separate set of dialogue histories, $H_\text{Eval}$, which is disjoint from the set of dialogue histories, $H_\text{Train}$ used at training time.
This ensures that the policy is not \textit{overfitting} our finite set of dialogue histories.
For each hyper-parameter combination, we train the policy between $400$ and $600$ episodes.
We select the policy which performs best w.r.t.\@ average return.
To keep notation brief, we call this policy \emph{Q-learning AMT}.

\subsection{Preliminary Evaluation}

In this section, we carry out a preliminary evaluation of the response model selection policies.

\begin{table}[t]
  \caption{Policy evaluation on AMT w.r.t.\@ score mean and score standard deviation (std). $90\%$ confidence intervals are given for means (after $\pm$) and standard deviations (in square brackets).} \label{tabel:policy_amt_evaluation}
  \small
  \centering
    \begin{tabular}{lcccc}
     \toprule
     & \multicolumn{2}{c}{\textbf{Full test set}} & \multicolumn{2}{c}{\textbf{Difficult test set}} \\ 
     \textbf{Policy} & \textbf{Score mean} & \textbf{Score std} & \textbf{Score mean} & \textbf{Score std} \\
    \midrule
    \emph{Alicebot} & $2.19 \pm 0.03$ & $1.17$ $[1.15, 1.20]$ & $1.79 \pm 0.03$ & $0.88$ $[0.86, 0.90]$ \\
    \emph{Evibot + Alicebot} & $2.25 \pm 0.04$ & $1.22$ $[1.20, 1.25]$ & $1.79 \pm 0.03$ & $0.86$ $[0.84, 0.88]$ \\
    \emph{Supervised AMT} & $2.63 \pm 0.04$ & $1.34$ $[1.31, 1.37]$ & $2.34 \pm 0.04$ & $1.26$ $[1.23, 1.29]$ \\
    \emph{Off-policy REINFORCE} & $2.61 \pm 0.04$ & $1.33$ $[1.31, 1.36]$ & $2.30 \pm 0.04$ & $1.25$ $[1.22, 1.28]$ \\
    \emph{Q-learning AMT} & $\mathbf{2.64 \pm 0.04}$ & $\mathbf{1.37}$ $\mathbf{[1.34, 1.40]}$ & $\mathbf{2.35 \pm 0.04}$ & $\mathbf{1.31}$ $\mathbf{[1.28, 1.34]}$ \\ \bottomrule
    \end{tabular}
\end{table}

\textbf{AMT Evaluation}: We first evaluate the learned policies on the w.r.t.\@ the human scores in the AMT test set.
We measure the average performance as a real-valued scalar, where the label $\textit{"Very poor"}$ is given a score of $1$, label $\textit{"Poor"}$ is given a score of $2$ and so on.
We also report standard deviations for the scores, which measure the \textit{variance} or \textit{risk} the policies are willing to take; higher standard deviations indicate that a policy is more likely to select responses which result in extreme labels (e.g.\@ \textit{"Very poor"} and \textit{"Excellent"}).
For both means and standard deviations we report $90\%$ confidence intervals estimated under the assumption that the scores are Gaussian-distributed.
In addition to measuring performance on the full test set, we also measure performance on a subset of the test set where neither \emph{Alicebot} nor \emph{Evibot} had responses labeled \textit{"Good"} or \textit{"Excellent"}.
These are test examples, where an appropriate response is likely to come only from some of the other models.
Determining an appropriate response for these examples is likely to be more difficult.
We refer to this subset as the \textit{"Difficult test set"}.

We evaluate the policies \emph{Supervised AMT}, \emph{Off-policy REINFORCE} and \emph{Q-learning AMT}. In addition, we also evaluate two heuristic policies: 1) a policy selecting only \emph{Alicebot} responses called \emph{Alicebot}, and 2) a policy selecting \emph{Evibot} responses when possible and \emph{Alicebot} responses otherwise, called \emph{Evibot + Alicebot}.

The results are given in Table \ref{tabel:policy_amt_evaluation}.
The results show that the three learned policies are all significantly better w.r.t.\@ mean score compared to both \emph{Alicebot} and \emph{Evibot + Alicebot}.
Not surprisingly, this difference is amplified on the difficult test set.
\emph{Q-learning AMT}, \emph{Supervised AMT} and \emph{Off-policy REINFORCE} appear to perform overall equally well.
This shows that machine learning has helped learn effective policies, able to select other model responses when neither the \emph{Alicebot} and \emph{Evibot} responses are appropriate.
Next, the results show that \emph{Q-learning AMT} has higher standard deviations than the other policies on both the full test set and the difficult test set.
Furthermore, since these standard deviations are evaluated at the level of a single response, we might expect this variability to compound throughout an entire conversation.
This strongly indicates that \emph{Q-learning AMT} is more risk tolerant than the other policies.

\begin{table}[ht]
  \caption{Off-policy evaluation w.r.t.\@ expected (average) Alexa user score and number of time steps (excluding priority responses) on test set.} \label{tabel:offpolicy_evaluation}
  \small
  \centering 
    \begin{tabular}{lcc}
     \toprule
     \textbf{Policy} & \textbf{Alexa user score} & \textbf{Time steps} \\
    \midrule
    \emph{Supervised AMT} & $2.06$ & $8.19$ \\
    \emph{Supervised Learned Reward} & $0.94$ & $3.66$ \\
    \emph{Off-policy REINFORCE} & $\mathbf{2.45}$ & $\mathbf{10.08}$ \\
    \emph{Off-policy REINFORCE Learned Reward} & $1.29$ & $5.02$ \\
    \emph{Q-learning AMT} & $2.08$ & $8.28$ \\ \bottomrule
    \end{tabular}
\end{table}

\textbf{Off-policy Evaluation}: One way to evaluate the selection policies is by using the off-policy evaluation given in eq.\@ \eqref{eq:offpolicy_reinforce_evaluation}.
This equation provides an estimate of the expected Alexa user score under each policy.\footnote{For the policies parametrized as action-value functions, we transform eq.\@ \eqref{eq:action_value_function} to eq.\@ \eqref{eq:stochastic_policy} by setting $f_\theta = Q_\theta$ and fitting the temperature parameter $\lambda$ on the \emph{Off-policy REINFORCE} development set.}
As described earlier, the same equation can be used to estimate the expected number of time steps per episode (excluding priority responses).

The expected (average) Alexa user score and number of time steps per episode (excluding priority responses) are given in Table \ref{tabel:offpolicy_evaluation}.
Here we observe that the \emph{Off-policy REINFORCE} performs best followed by \emph{Q-learning AMT} and \emph{Supervised AMT} w.r.t.\@ expected Alexa user score.
\emph{Off-policy REINFORCE} reaches $2.45$, which is a major $17.8\%$ improvement over the second best performing model \emph{Q-learning AMT}.
However, this advantage should be taken with a grain of salt.
As discussed earlier, the off-policy evaluation in eq.\@ \eqref{eq:offpolicy_reinforce_evaluation} is a biased estimator since the importance weights have been truncated.
Moreover, \emph{Off-policy REINFORCE} has been trained specifically to maximize this biased estimator, while all other policies have been trained to maximize other objective functions.
Similarly, w.r.t.\@ expected number of time steps, \emph{Off-policy REINFORCE} reaches the highest number of time steps followed by \emph{Q-learning AMT} and \emph{Supervised AMT}.
As before, we should take this result with a grain of salt, since this evaluation is also biased and does not take into account priority responses.
Further, it's not clear that increasing the number of time steps will increase user scores.
Nevertheless, \emph{Off-policy REINFORCE}, \emph{Q-learning AMT} and \emph{Supervised AMT} appear to be our prime candidates for further experiments.


\begin{figure}[ht]
  \centering
  \includegraphics[scale=0.31]{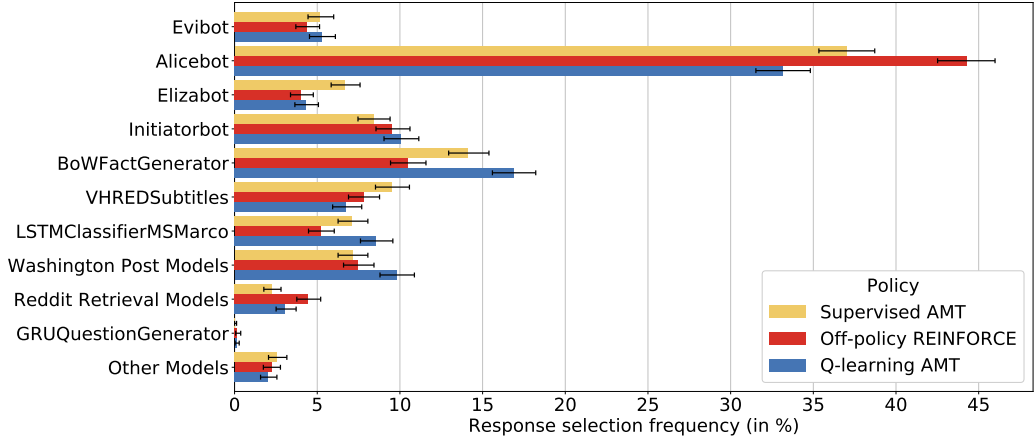}
  \caption{Response model selection probabilities across response models for \emph{Supervised AMT}, \emph{Off-policy REINFORCE} and \emph{Q-learning AMT} on the AMT label test dataset. 
  $95\%$ confidence intervals are shown based on the Wilson score interval for binomial distributions.}
  \label{fig:rl_response_model_selection_frequencies}
\end{figure}

\textbf{Response Model Selection Frequency}:
Figure \ref{fig:rl_response_model_selection_frequencies} shows the frequency with which \emph{Supervised AMT}, \emph{Off-policy REINFORCE} and \emph{Q-learning AMT} select different response models.
We observe that the policy learned using \emph{Off-policy REINFORCE} tends to strongly prefer \emph{Alicebot} responses over other models.
The \emph{Alicebot} responses are among the safest and most topic-dependent, generic responses in the system, which suggests that \emph{Off-policy REINFORCE} has learned a highly \textit{risk averse strategy}.
On the other hand, the \emph{Q-learning AMT} policy selects \emph{Alicebot} responses substantially less often than both \emph{Off-policy REINFORCE} and \emph{Supervised AMT}.
Instead, \emph{Q-learning AMT} tends to prefer responses retrieved from Washington Post and from Google search results.
These responses are semantically richer and have the potential to engage the user more deeply in a particular topic, but they are also more risky (e.g.\@ a bad choice could derail the entire conversation.).
This suggests that \emph{Q-learning AMT} has learned a more \textit{risk tolerant strategy}.
One possible explanation for this difference is that \emph{Q-learning AMT} was trained using simulations.
By learning online from simulations, the policy has been able to explore new actions and discover high-level strategies lasting multiple time steps.
In particular, the policy has been allowed to experiment with riskier actions and to learn \textit{remediation} or \textit{fall-back strategies}, in order to handle cases where a risky action fails.
This might also explain its stronger preference for \emph{BoWFactGenerator} responses, which might be serving as a fall-back strategy by outputting factual statements on the current topic.
This would have been difficult to learn for \emph{Off-policy REINFORCE}, since the sequence of actions for such high-level strategies are sparsely observed in the data and, when they are observed, the corresponding returns (Alexa user scores) have high variance.

A second observation is that \emph{Q-learning AMT} has the strongest preference for \emph{Initiatorbot} among the three policies.
This could indicate that \emph{Q-learning AMT} leans towards a \textit{system-initiative strategy} (e.g.\@ a strategy where the system tries to maintain control of the conversation by asking questions, changing topics and so on).
Further analysis is needed to confirm this.

\begin{table}[t]
  \caption{Policy evaluation using the \emph{Abstract Discourse MDP} w.r.t.\@ average return, average reward per time step and average episode length on dev set ($\pm$ standard deviations). The reward function is based on \emph{Supervised AMT}.} \label{tabel:mdp_evaluation}
  \small
  \centering
    \begin{tabular}{lccc@{\hskip 0.15in}ccccc}
     \toprule
     \textbf{Policy} & \textbf{Average return} & \textbf{Average reward per time step} & \textbf{Average dialogue length} \\
    \midrule
    \emph{Random} & $-32.18 \pm 31.77$ & $-0.87 \pm 0.24$ & $34.29 \pm 33.02$ \\
    \emph{Alicebot} & $-15.56 \pm 15.61$ & $-0.37 \pm 0.16$ & $42.01 \pm 42.00$  \\
    \emph{Evibot + Alicebot} & $-11.33 \pm 12.43$ & $-0.29 \pm 0.19$ & $37.5 \pm 38.69$  \\
    \emph{Supervised AMT} & $\mathbf{-6.46 \pm 8.01}$ & $\mathbf{-0.15 \pm 0.16}$ & $\mathbf{42.84 \pm 42.92}$  \\
    \emph{Supervised Learned Reward} & $-24.19 \pm 23.30$ & $-0.73 \pm 0.27$ & $31.91 \pm 30.09$ \\
    
    \emph{Off-policy REINFORCE} & $\mathbf{-7.30 \pm 8.90}$ & $\mathbf{-0.16 \pm 0.16}$ & $\mathbf{43.24 \pm 43.58}$ \\
    \parbox[c][2.65em][c]{0.225\textwidth}{\emph{Off-policy REINFORCE} \\ \protect{\hphantom{\ }} \emph{Learned Reward}} & $-10.19 \pm 11.15$ & $-0.28 \pm 0.19$ & $35.51 \pm 35.05$ \\
    \emph{Q-learning AMT} & $\mathbf{-6.54 \pm 8.02}$ & $\mathbf{-0.15 \pm 0.18}$ & $\mathbf{40.68 \pm 39.13}$ \\ \bottomrule
    \end{tabular}
\end{table}

\textbf{Abstract Discourse MDP Evaluation}
Next, we can evaluate the performance of each policy w.r.t.\@ simulations in the \emph{Abstract Discourse MDP}.
We simulate 500 episodes under each policy and evaluate it w.r.t.\@ average return, average reward per time step and dialogue length.
In addition to evaluating the five policies described earlier, we also evaluate three heuristic policies: 1) a policy selecting responses at random called \emph{Random}, 2) the \emph{Alicebot} policy, and 3) the \emph{Evibot + Alicebot} policy.
Evaluating these models will serve to validate the approximate MDP.

The results are given in Table \ref{tabel:mdp_evaluation}.
We observe that \emph{Supervised AMT} performs best w.r.t.\@ average return and average reward per time step.
However, this comes as no surprise.
The reward function in the MDP is defined as \emph{Supervised AMT}, so by construction this policy achieves the highest reward per time step.
Next we observe that \emph{Q-learning AMT} is on par with \emph{Supervised AMT}, both achieving same $-0.15$ average reward per time step.
Second in line comes \emph{Off-policy REINFORCE}, achieving an average reward per time step of $-0.16$.
However, \emph{Off-policy REINFORCE} also achieved the highest average dialogue length of $43.24$.
At the other end of the spectrum comes, as expected, the \emph{Random} policy performing worst w.r.t.\@ all metrics.
In comparison, both \emph{Alicebot} and \emph{Evibot + Alicebot} perform better w.r.t.\@ all metrics, with \emph{Evibot + Alicebot} achieving the best average return and average reward per time step out of the three heuristic policies.
This validates the utility of the \emph{Abstract Discourse MDP} as an environment for training and evaluating policies.
Overall, \emph{Off-policy REINFORCE}, \emph{Q-learning AMT} and \emph{Supervised AMT} still appear to be the best performing models in the preliminary evaluation.

\begin{figure}[H]
  \centering
  \includegraphics[scale=0.375]{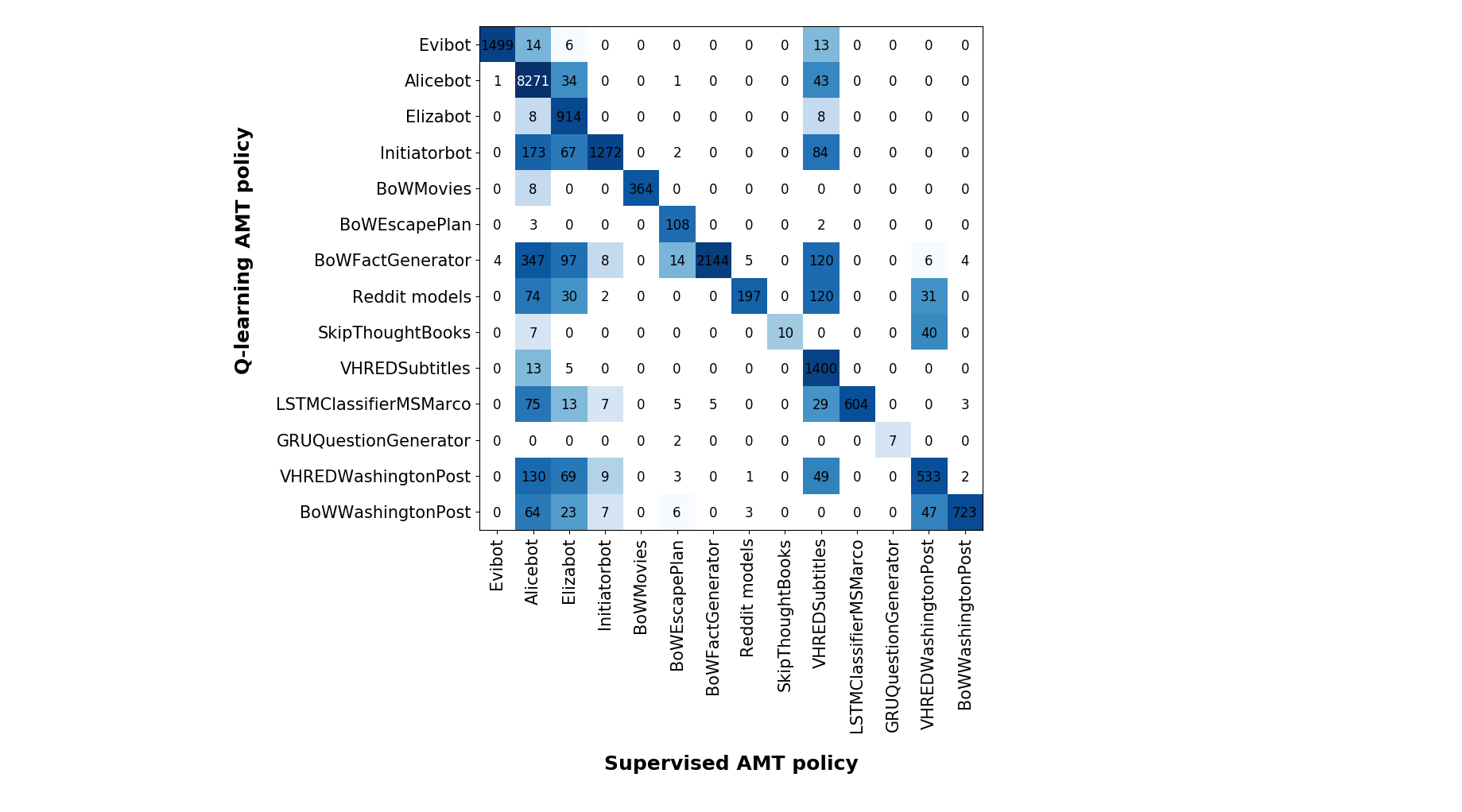}
  \caption{Contingency table comparing selected response models between \emph{Supervised AMT} and \emph{Q-learning AMT}. The cells in the matrix show the number of times the \emph{Supervised AMT} policy selected the row response model and the \emph{Q-learning AMT} policy selected the column response model. The cell frequencies were computed by simulating 500 episodes under the Q-learning policy in the \emph{Abstract Discourse MDP}. Note that all models retrieving responses from Reddit have been agglomerated into the class \emph{Reddit models}.}
  \label{fig:rl_response_model_selection_frequencies_q_learning_supervised_policy}
\end{figure}

Finally, we compare \emph{Q-learning AMT} with \emph{Supervised AMT} w.r.t.\@ the action taken in states from episodes simulated in the \emph{Abstract Discourse MDP}.
As shown in Figure \ref{fig:rl_response_model_selection_frequencies_q_learning_supervised_policy},
the two policies diverge w.r.t.\@ several response models.
When \emph{Supervised AMT} would have selected topic-independent, generic \emph{Alicebot} and \emph{Elizabot} responses, \emph{Q-learning AMT} often selects \emph{BoWFactGenerator}, \emph{Initiatorbot} and \emph{VHREDWashingtonPost} responses.
For example, there were 347 instances where \emph{Supervised AMT} selected \emph{Alicebot}, but where \emph{Q-learning AMT} selected \emph{BoWFactGenerator}.
Similarly, where \emph{Supervised AMT} would have preferred generic \emph{VHREDSubtitle} responses, \emph{Q-learning AMT} often selects responses from \emph{BoWFactGenerator}, \emph{InitiatorBot} and \emph{VHREDRedditSports}.
This supports our previous analysis showing that \emph{Q-learning AMT} has learned a more risk tolerant strategy, which involves response models with semantically richer content.

In the next section, we evaluate these policies with real-world users.

\section{A/B Testing Experiments} \label{sec:ab_testing_experiments}
To evaluate the dialogue manager policies described in the previous section, we carry out A/B testing experiments.
During each A/B testing experiment, we evaluate several policies for selecting the response model.
When Alexa users start a conversation with the system, they are automatically assigned to a random policy and afterwards their dialogues and final scores are recorded.

A/B testing allows us to accurately compare different dialogue manager policies by keeping all other system factors constant (or almost constant).
This is in contrast to evaluating the system performance over time, when the system is continuously being modified.
In such a situation, it is often difficult to evaluate the improvement or degradation of performance w.r.t.\@ particular system modifications.

However, even during our A/B testing experiments, the distribution over Alexa users still changes through time.
Different types of users will be using the system depending on the time of day, weekday and holiday season.
In addition, the user expectations towards our system change over time as they interact with other socialbots in the competition.
In other words, we must consider the Alexa user distribution as following a non-stationary stochastic process.
Therefore, we take two steps to reduce confounding factors and correlations between users.
First, during each A/B testing experiment, we evaluate all policies of interest simultaneously.
This ensures that we have approximately the same number of users interacting with each policy w.r.t.\@ time of day and weekday.
This minimizes the effect of changes in the user distribution on the final user scores \textit{within} that period.
However, since the user distribution changes between the A/B testing experiments, we still cannot accurately compare policy performance \textit{across} A/B testing experiments.
Second, we discard scores from returning users (i.e.\@ users who have already evaluated the system once).
Users who are returning to the system are likely to be influenced by their previous interactions with the system.
For example, users who previously had a positive experience with the system may be biased towards giving high scores in their next interaction.
Further, the users who return to the system are likely to belong to a particular subpopulation of users.
This particular group of users may inherently have more free time and be more willing to engage with socialbots than other users.
Discarding returning user scores ensures that the evaluation is not biased towards this subpopulation of users.
By discarding scores from returning users, we also ensure that the evaluation counts every user exactly once.
Finally, it should be noted that we ignore dialogues where the Alexa user did not give a score.
This inevitably biases our evaluation, since users who do not provide a score are likely to have been dissatisfied with the system or to have been expecting different functionality (e.g.\@ \textit{non-conversational activities}, such as playing music, playing games or taking quizzes).
One potential remedy is to have all dialogues evaluated by a third-party (e.g.\@ by asking human annotators on Amazon Mechanical Turk to evaluate the dialogue), but that is beyond the scope of these experiments.


\subsection{A/B Testing Experiment \#1}
The first A/B testing experiment was carried out between July 29th, 2017 and August 6th, 2017.
We tested six dialogue manager policies: \emph{Evibot + Alicebot}, \emph{Supervised AMT}, \emph{Supervised Learned Reward}, \emph{Off-policy REINFORCE}, \emph{Off-policy REINFORCE Learned Reward} and \emph{Q-learning AMT}.
For \emph{Off-policy REINFORCE} and \emph{Off-policy REINFORCE Learned Reward}, we use the greedy variant defined in eq.\@ \eqref{eq:greedy_stochastic_policy}.

This experiment occurred early in the Amazon Alexa Prize competition.
This means that Alexa users have few expectations towards our system (e.g.\@ expectations that the system can converse on a particular topic, or that the system can engage in \textit{non-conversational activities}, such as playing word games or taking quizzes).
Further, the period July 29th - August 6th overlaps with the summer holidays in the United States.
This means that we might expect more children to interact with system than during other seasons.

\begin{table}[t]
  \caption{First A/B testing experiment with six different policies ($\pm$ 95\% confidence intervals). Star $^{*}$ indicates policy is significantly better than other policies at $95\%$ statistical significance level.} \label{tabel:ab_testing_round_one}
  \small
  \centering
    \begin{tabular}{lcccc}
     \toprule
     \textbf{Policy} & \textbf{User score} & \textbf{Dialogue length} & \textbf{Pos.\@ utterances} & \textbf{Neg.\@ utterances} \\
    \midrule
    \emph{Evibot + Alicebot} & $2.86 \pm 0.22 $ & $ 31.84 \pm 6.02 $ & $2.80\% \pm 0.79 $ & $5.63\% \pm 1.27$ \\
    \emph{Supervised AMT} & $ 2.80 \pm 0.21 $ & $ 34.94 \pm 8.07 $ & $\mathbf{4.00\% \pm 1.05}$ & $8.06\% \pm 1.38$ \\
    \emph{Supervised Learned Reward} & $2.74 \pm 0.21 $ & $ 27.83 \pm 5.05 $ & $2.56\% \pm 0.70 $ & $6.46\% \pm 1.29$ \\
    \emph{Off-policy REINFORCE} & $2.86 \pm 0.21 $ & $\mathbf{37.51 \pm 7.21}$ & $3.98\% \pm 0.80 $ & $6.25 \pm 1.28$ \\
    \parbox[c][2.65em][c]{0.225\textwidth}{\emph{Off-policy REINFORCE} \\ \protect{\hphantom{\ }} \emph{Learned Reward}} & $2.84 \pm 0.23 $ & $ 34.56 \pm 11.55 $ & $ 2.79\% \pm 0.76$ & $ 6.90\% \pm 1.45$ \\
    \emph{Q-learning AMT}* & $\mathbf{3.15 \pm 0.20}$ & $ 30.26 \pm 4.64 $ & $3.75\% \pm 0.93 $ & $\mathbf{5.41\% \pm 1.16}$ \\ \bottomrule
    \end{tabular}
\end{table}

\begin{table}[t]
  \caption{Amazon Alexa Prize semi-finals average team statistics provided by Amazon.} \label{tabel:amazon_alexa_team_stats}
  \small
  \centering
    \begin{tabular}{lcc}
     \toprule
     \textbf{Policy} & \textbf{User score} & \textbf{Dialogue length} \\
    \midrule
    \emph{All teams} & $2.92$ & $22$ \\
    \emph{Non-finalist teams} & $2.81$ & $22$ \\
    \emph{Finalist teams} & $\mathbf{3.31}$ & $\mathbf{26}$ \\ \bottomrule
    \end{tabular}
\end{table}

\textbf{Policy Evaluation}
The results are given in Table \ref{tabel:ab_testing_round_one}.\footnote{$95\%$ confidence intervals are computed under the assumption that the Alexa user scores for each policy are drawn from a Gaussian distribution with its own mean and variance. This is an approximation, since the Alexa user scores only have support on the interval $[1, 5]$.}
The table shows the average Alexa user scores, average dialogue length, average percentage of positive user utterances and average percentage of negative user utterances. 
In total, over a thousand user ratings were collected after discarding returning users. Ratings were collected after the end of the semi-finals competition, where all ratings had been transcribed by human annotators.
Each policy was evaluated by about two hundred unique Alexa users.

As expected from our preliminary evaluation, we observe that \emph{Q-learning AMT} and \emph{Off-policy REINFORCE} perform best among all policies w.r.t.\@ user scores.
\emph{Q-learning AMT} obtained an average user score of $3.15$, which is significantly higher than all other policies at a $95\%$ statistical significance level w.r.t.\@ a one-tailed two-sample t-test.
In comparison, the average user score for all the teams in the competition during the semi-finals was only $2.92$.
Interestingly, \emph{Off-policy REINFORCE} achieved the longest dialogues with an average length of $37.51$.
This suggests \emph{Off-policy REINFORCE} yields highly engaging conversations.
In comparison, in the semi-finals, the average dialogue length of all teams was $22$ and of the finalist teams was $26$.
We also observe that \emph{Off-policy REINFORCE} had a slightly higher percentage of user utterances with negative sentiment compared to \emph{Q-learning AMT}.
This potentially indicates that the longer dialogues also include some frustrated interactions (e.g.\@ users who repeat the same questions or statements in the hope that the system will return a more interesting response next time).
The remaining policies achieved average Alexa user scores between $2.74$ and $2.86$, with the heuristic policy \emph{Evibot + Alicebot} obtaining $2.86$.
This suggests that the other policies have not learned to select responses more appropriately than the \emph{Evibot + Alicebot} heuristic.

In conclusion, the results indicate that the \textit{risk tolerant} learned by the \emph{Q-learning AMT} policy performs best among all policies.
This shows that learning a policy through simulations in an \emph{Abstract Discourse MDP} may serve as a fruitful path towards developing open-domain socialbots.
In addition, the performance of \emph{Off-policy REINFORCE} indicates that optimizing the policy directly towards Alexa user scores could also potentially yield improvements.
However, further investigation is required.

\textbf{Length Analysis}

In an effort to further understand how the policies differ from each other, we carry out an analysis of the policies performance as a function of dialogue length.
Although, we have recorded only a limited amount of data for dialogues with a particular length, this analysis could help illuminate directions for future experiments.

Table \ref{tabel:ab_testing_round_one_varying_length} shows the average Alexa user scores w.r.t.\@ four dialogue length intervals for the six policies.
The estimates are based on between 30-70 Alexa user ratings for each policy and interval combination.
First, we observe that \emph{Q-learning AMT} performs better than all other policies for all intervals except the medium-short interval ($10-19$, or $5-10$ back-and-forth turns).
Further, its high performance for the long intervals ($20 - 39$ and $\geq 40$) would suggest that \emph{Q-learning AMT} performs excellent in long dialogues.
The other learned policies \emph{Supervised AMT}, \emph{Off-policy REINFORCE} and \emph{Off-policy REINFORCE Learned Reward} also appear to perform excellent in long dialogues.
On the other hand, the heuristic \emph{Evibot + Alicebot} policy  and the \emph{Supervised Learned Reward} policy appear to perform poorly in long dialogues, but that is not surprising given their low overall performance.
In particular, \emph{Supervised Learned Reward} seems to be performing well only for very short dialogues.
This potentially indicates that the policy fails to either maintain user engagement or memorize longer-term context.
However, further investigation is required.

\begin{table}[t]
  \caption{First A/B testing experiment user scores with six different policies w.r.t.\@ varying dialogue length ($\pm$ one standard deviation).} \label{tabel:ab_testing_round_one_varying_length}
  \small
  \centering
    \begin{tabular}{lcccc}
     \toprule
     & \multicolumn{4}{c}{\textbf{Dialogue length}}\\ 
     \textbf{Policy} & \textbf{< 10} & \textbf{10 - 19} & \textbf{20 - 39} & \textbf{$\geq$ 40} \\
    \midrule
    \emph{Evibot + Alicebot} & $2.88 \pm 1.71$ & $2.58 \pm 1.33$ & $2.93 \pm 1.28$ & $2.99 \pm 1.37$ \\
    \emph{Supervised AMT} & $2.91 \pm 1.59$ & $2.64 \pm 1.38$ & $2.60 \pm 1.40$ & $3.13 \pm 1.43$ \\
    \emph{Supervised Learned Reward} & $3.31 \pm 1.43$ & $2.45 \pm 1.57$ & $2.19 \pm 1.38$ & $2.90 \pm 1.54$ \\
    \emph{Off-policy REINFORCE} & $2.99 \pm 1.64$ & $\mathbf{2.72 \pm 1.57}$ & $2.56 \pm 1.31$ & $3.26 \pm 1.45$ \\
    \parbox[c][2.65em][c]{0.225\textwidth}{\emph{Off-policy REINFORCE} \\ \protect{\hphantom{\ }} \emph{Learned Reward}} & $2.91 \pm 1.64$ & $2.53 \pm 1.45$ & $2.9 \pm 1.56$ & $3.14 \pm 1.36$ \\
    \emph{Q-learning AMT} & $\mathbf{3.46 \pm 1.40}$ & $2.60 \pm 1.45$ & $\mathbf{3.19 \pm 1.39}$ & $\mathbf{3.31 \pm 1.33}$ \\ \bottomrule    \end{tabular}
\end{table}

\textbf{Topical Specificity and Coherence}

We carry out an analysis of the topical specificity and coherence of the different policies.
This analysis aims to quantify how much each policy stays on topic (e.g.\@ whether the policy selects responses on the current topic or on new topics) and how specific its content is (e.g.\@ how frequently the policy selects generic, topic-independent responses).
This analysis is carried out at the utterance level, where we are fortunate to have more recorded data.

The results are shown in Table \ref{tabel:ab_testing_round_one_topical_analysis}.
For topic specificity, we measure the average number of noun phrases per user utterance and the average number of noun phrases per system utterance.\footnote{We use \url{https://spacy.io} version 1.9.0 to detect noun phrases with the package \textit{"en\_core\_web\_md-1.2.1"}.}
The more topic specific the user is, the higher we would expect the first metric to be.
Similarly, the more topic specific the system is the higher we would expect the second metric to be.
For topic coherence, we measure the word overlap between the user's utterance and the system's response, as well as word overlap between the user's utterance and the system's response at the next turn.
The more the policy prefers to stay on topic, the higher we would expect these two metrics to be.

As shown in the table, \emph{Q-learning AMT} has obtained significantly higher scores w.r.t.\@ both word overlap metrics and the average number of noun phrases per system utterance.
This indicates that the \emph{Q-learning AMT} policy has the highest topical coherency among all six policies, and that it generates the most topic specific (semantically rich) responses.
This is in line with our previous analysis, where we found that \emph{Q-learning} follows a highly \textit{risk tolerant} strategy.
Next in line, comes \emph{Supervised AMT}, which also appears to maintain high topic specificity and coherence.
In fact, \emph{Supervised AMT} obtained the highest metric w.r.t.\@ number of noun phrases per user utterance, which indicates that this policy is encouraging the user to give more topic specific responses.
Afterwards comes \emph{Off-policy REINFORCE} and \emph{Off-policy REINFORCE Learned Reward}, which tend to select responses with significantly less noun phrases and less word overlap.
This is also in line with our previous analysis, where we found that \emph{Off-policy REINFORCE} follows a \textit{risk averse} strategy.
Finally, the heuristic policy \emph{Evibot + Alicebot} selects responses with very few noun phrases and least word overlap among all policies.
This indicates that the heuristic policy might be the least topic coherent policy, and that it mainly selects generic, topic-independent responses.

\begin{table}[t]
  \caption{First A/B testing experiment topical specificity and coherence of the six different policies. The columns are average number of noun phrases per user utterance (User NPs), average number of noun phrases per system utterance (System NPs), average number of overlapping words between the user's utterance and the system's response (Word overlap $t \to t + 1$), and average number of overlapping words between the user's utterance and the system's response in the next turn (Word overlap $t \to t + 3$).
  $95\%$ confidence intervals are also shown. Stop words are excluded.} \label{tabel:ab_testing_round_one_topical_analysis}
  \small
  \centering
    \begin{tabular}{lcccc}
     \toprule
     \textbf{Policy} & \textbf{User NPs} & \textbf{System NPs} & \textbf{Word overlap} & \textbf{Word overlap} \\
     & & & \textbf{$t \to t+1$} & \textbf{$t \to t+3$} \\ \midrule
    \emph{Evibot + Alicebot} & $0.55 \pm 0.03$ & $1.05 \pm 0.05$ & $7.33 \pm 0.21$ & $7.31 \pm 0.22$ \\
    \emph{Supervised AMT} & $\mathbf{0.62 \pm 0.03}$ & $1.75 \pm 0.07$ & $10.48 \pm 0.28$ & $10.65 \pm 0.29$ \\
    \emph{Supervised Learned Reward} & $0.57 \pm 0.03$ & $1.50 \pm 0.07$ & $8.35 \pm 0.29$ & $8.36 \pm 0.31$ \\
    \emph{Off-policy REINFORCE} & $0.59 \pm 0.02$ & $1.45 \pm 0.05$ & $9.05 \pm 0.21$ & $9.14 \pm 0.22$ \\
    \parbox[c][2.65em][c]{0.225\textwidth}{\emph{Off-policy REINFORCE} \\ \protect{\hphantom{\ }} \emph{Learned Reward}} & $0.61 \pm 0.03$ & $1.04 \pm 0.06$ & $7.42 \pm 0.25$ & $7.42 \pm 0.26$ \\
    \emph{Q-learning AMT} & $0.58 \pm 0.03$ & $\mathbf{1.98 \pm 0.08}$ & $\mathbf{11.28 \pm 0.30}$ & $\mathbf{11.52 \pm 0.32}$ \\ \bottomrule    \end{tabular}
\end{table}

\textbf{Initiatorbot Evaluation}
This experiment also allowed us to analyze the outcomes of different conversation starter phrases given by the \emph{Initiatorbot}.
We carried out this analysis by computing the average Alexa user score for each of the 40 possible phrases.
We found that phrases related to news (e.g.\@ \textit{"Do you follow the news?"}), politics (e.g.\@ \textit{"Do you want to talk about politics?"}) and travelling (e.g.\@ \textit{"Tell me, where do you like to go on vacation?"}) performed poorly across all policies.
On the other hand, phrases related to animals (e.g.\@ \textit{"Do you have pets?"} and \textit{"What is the cutest animal you can think of?"}), movies (e.g.\@ \textit{"Let's talk about movies. What's the last movie you watched?"}) and food (e.g.\@ \textit{"Let's talk about food. What is your favorite food?"}) performed well across all policies.
For example, conversations where the \emph{Initiatorbot} asked questions related to news and politics had an average Alexa user score of only $2.91$ for the top two systems (\emph{Off-policy REINFORCE} and \emph{Q-learning AMT}).
Mean while, conversations where the \emph{Initiatorbot} asked questions about animals, food and movies the corresponding average Alexa user score was $3.17$.
We expected the conversation topic to affect user engagement, however it is surprising that these particular topics (animals, food and movies) were the most preferred ones.
One possible explanation is that our system does not perform well on news, politics and travelling topics.
However, the system already had several response models dedicated to discussing news and politics: six sequence-to-sequence models extracting responses from Reddit news and Reddit politics, two models extracting responses from Washington Post user comments and the \emph{BoWTrump} model extracting responses from Donald J. Trump's Twitter profile.
In addition, \emph{Evibot} is capable of answering many factual questions about news and politics and \emph{BoWFactGenerator} contains hundreds of facts related to news and politics.
As such, there may be another more plausible explanation for users' preferences towards topics, such as animals, movies and food.
One likely explanation is the age group of the users.
While inspecting our conversational transcripts, we observed that many users interacting with the system appeared to be children or teenagers.
It would hardly come as a surprise if this user population would prefer to talk about animals, movies and foods rather than news, politics and travels.

\subsection{A/B Testing Experiment \#2}

The second A/B testing experiment was carried out between August 6th, 2017 and August 15th, 2017.
We tested two dialogue manager policies: \emph{Off-policy REINFORCE} and \emph{Q-learning AMT}.
As before, we use the greedy variant of \emph{Off-policy REINFORCE} defined in eq.\@ \eqref{eq:greedy_stochastic_policy}.

This experiment occurred at the end of the Amazon Alexa Prize competition semi-finals.
This means that many Alexa users have already interacted with other socialbots in the competition, and therefore are likely to have developed expectations towards the systems.
These expectations are likely to involve conversing on a particular topic or engaging in \textit{non-conversational activities}, such as playing games).
Further, the period August 6th - August 15th overlaps with the end of the summer holidays and the beginning of the school year in the United States.
This means that we should expect less children to interact with the system than in the previous A/B testing experiment.

\begin{table}[t]
  \caption{Second A/B testing experiment with two different policies ($\pm$ 95\% confidence intervals).} \label{tabel:ab_testing_round_two}
  \small
  \centering
    \begin{tabular}{lcccc}
     \toprule
     \textbf{Policy} & \textbf{User score} & \textbf{Dialogue length} & \textbf{Pos.\@ utterances} & \textbf{Neg.\@ utterances} \\
    \midrule
    \emph{Off-policy REINFORCE} & $\mathbf{3.06 \pm 0.12}$ & $\mathbf{34.45 \pm 3.76}$ & $3.23\% \pm 0.45$ & $7.97\% \pm 0.85$ \\
    \emph{Q-learning AMT} & $2.92 \pm 0.12$ & $31.84 \pm 3.69$ & $\mathbf{3.38\% \pm 0.50}$ & $\mathbf{7.61\% \pm 0.84}$
    \\ \bottomrule
    \end{tabular}
\end{table}

\textbf{Policy Evaluation}
The results are given in Table \ref{tabel:ab_testing_round_two}.
In total, about eight hundred user ratings were collected after discarding returning users.
As such, each policy was evaluated by about six hundred unique Alexa users.
As before, all ratings were transcribed by human annotators.

We observe that both \emph{Off-policy REINFORCE} and \emph{Q-learning AMT} perform better than the policies in the previous experiment.
However, in this experiment, \emph{Off-policy REINFORCE} achieved an average Alexa user score of $3.06$ while \emph{Q-learning AMT} achieved a lower score of only $2.92$.
Nonetheless, \emph{Off-policy REINFORCE} is not statistically significantly better.
In this experiment, there is also no significant difference between the two policies w.r.t.\@ percentage of positive and negative user utterances.

As discussed earlier, the performance difference compared to the previous A/B testing experiment could be due to the change in user profiles and user expectations.
At this point in time, more of the Alexa users have interacted with socialbots from other teams.
Mean while, all socialbots have been evolving.
Therefore, user expectations towards our system are likely to be higher now.
Further, since the summer holidays have ended, less children and more adults are expected to interact with our system.
It is plausible that these adults also have higher expectations towards the system, and even more likely that they are less playful and less tolerant towards mistakes.
Given this change in user profiles and expectations, the \textit{risk tolerant} strategy learned by the \emph{Q-learning AMT} policy is likely to fare poorly compared to the \textit{risk averse} strategy learned by \emph{Off-policy REINFORCE}.

\subsection{A/B Testing Experiment \#3}

The third A/B testing experiment was carried out between August 15th, 2017 and August 21st, 2017.
Due to the surprising results in the previous A/B testing experiment, we decided to continue testing the two dialogue manager policies \emph{Off-policy REINFORCE} and \emph{Q-learning AMT}.
As before, we use the greedy variant of \emph{Off-policy REINFORCE} defined in eq.\@ \eqref{eq:greedy_stochastic_policy}.

This experiment occurred after the end of the Amazon Alexa Prize competition semi-finals.
As discussed before, this means that it is likely that  many Alexa users have already developed expectations towards the systems.
Further, the period August 15th - August 21st lies entirely within the beginning of the school year in the United States.
This means that we should expect less children to interact with the system than in the previous A/B testing experiment.

\textbf{Policy Evaluation}
The results are given in Table \ref{tabel:ab_testing_round_three}.
In total, about six hundred user ratings were collected after discarding returning users.
As such, each policy was evaluated by about three hundred unique Alexa users.
Unlike the previous two experiments, due to the semi-finals having ended, these ratings were not transcribed by human annotators.

\begin{table}[t]
  \caption{Third A/B testing experiment with two different policies ($\pm$ 95\% confidence intervals).} \label{tabel:ab_testing_round_three}
  \small
  \centering
    \begin{tabular}{lcccc}
     \toprule
     \textbf{Policy} & \textbf{User score} & \textbf{Dialogue length} & \textbf{Pos.\@ utterances} & \textbf{Neg.\@ utterances} \\
    \midrule
    \emph{Off-policy REINFORCE} & $3.03 \pm 0.18$ & $30.93 \pm 4.96$ & $2.72 \pm 0.59$ & $7.36 \pm 1.22$ \\
    \emph{Q-learning AMT} & $\mathbf{3.06 \pm 0.17}$ & $\mathbf{33.69 \pm 5.84}$ & $\mathbf{3.63 \pm 0.68}$ & $\mathbf{6.67 \pm 0.98}$     \\ \bottomrule
    \end{tabular}
\end{table}

We observe again that both \emph{Off-policy REINFORCE} and \emph{Q-learning AMT} perform better than the other policies evaluated in the first experiment.
However, in this experiment, \emph{Off-policy REINFORCE} only achieved an average Alexa user score of $3.03$ while \emph{Q-learning AMT} achieved the higher score of $3.06$.
As before, neither policy is statistically significantly better than the other.
Nevertheless, as in the first experiment, \emph{Q-learning AMT} achieved a higher percentage of positive utterances and a lower percentage of negative utterances than \emph{Off-policy REINFORCE}.
In this experiment, \emph{Q-learning AMT} also obtains the longest dialogues on average.
Overall, this experiment indicates that \emph{Q-learning AMT} is the better policy.

As before, the difference in performance compared to the previous A/B testing experiments is likely due to the change in user profiles and user expectations.
The fact that \emph{Q-learning AMT} now performs slightly better than \emph{Off-policy REINFORCE} might be explained by many different causes.
First, despite the confidence intervals and statistical tests presented earlier, it is of course possible that the previous A/B testing experiments did not have enough statistical power to accurately discriminate whether \emph{Q-learning AMT} or \emph{Off-policy REINFORCE} obtains the highest average user score.
Second, it is possible that the topics users want to discuss now are simply better handled by \emph{Q-learning AMT}.
Third, it is possible that adult users might only have a weak preference toward the risk averse  \emph{Q-learning AMT} policy, and that there is still a significant amount of children and teenagers interacting with the system even though the summer holidays have ended.
Finally, it is possible that the user population has grown tired of \emph{Off-policy REINFORCE}, which follows a risk averse strategy by responding with less semantic content.

\subsection{Discussion}
The two dialogue manager policies \emph{Q-learning AMT} and \emph{Off-policy REINFORCE} have demonstrated substantial improvements over all other policies, including policies learned using supervised learning and heuristic policies.
As discussed earlier, the \emph{Q-learning AMT} policy achieved an average Alexa user score substantially above the average score of all teams in the Amazon Alexa Prize competition semi-finals, without relying on non-conversational activities.
In addition, it also achieved a higher number of dialogue turns than both the average of all teams in the semi-finals and the average of all finalist teams in the semi-finals.
The policy \emph{Off-policy REINFORCE} similarly obtained a high number of dialogue, suggesting that the resulting conversations are far more engaging.
The results demonstrate the advantages of the overall ensemble approach, where many different models generate natural language responses and the dialogue manager policy selects one response among them.
The results also highlight the advantages of learning the policy using reinforcement learning techniques.
By optimizing the policy to maximize either real-world user scores or to maximize rewards in the \emph{Abstract Discourse MDP} (with a proxy reward function) we have demonstrated that significant gains can be achieved w.r.t.\@ both real-world user scores and number of dialogue turns.



\section{Related Work}

\textbf{Dialogue Manager Architecture}: Any open-domain conversational agent will have to utilize many different types of modules, such as modules for looking up information, modules for daily chitchat discussions, modules for discussing movies, and so on.
In this respect, our system architecture is related to some of the recent general-purpose dialogue system frameworks~\citep{zhao2016dialport,miller2017parlai,truong2017maca}.
These systems abstract away the individual modules into black boxes sharing the same interface, similar to the response models in our ensemble.
This, in turn, enables them to be controlled by an executive component (e.g.\@ a dialogue manager).

\textbf{Reinforcement Learning}:


Much work has applied reinforcement learning to training or improving dialogue systems. The idea that dialogue can be formulated as a sequential decision making problem based on a Markov decision process (MDP) appeared already in the 1990s for goal-oriented dialogue systems~\citep{singh1999reinforcement,singh2002optimizing,williams2007partially,young2013pomdp,paek2006reinforcement,henderson2008hybrid,pieraccini2009we,svgk15}.

One line of research in this area has focused on learning dialogue systems through simulations using abstract dialogue states and actions~\citep{eckert1997user,levin2000stochastic,chung2004developing,cuayahuitl2005human,georgila2006user,schatzmann2007agenda,heeman2009representing,traum2008multi,georgila2011reinforcement,lee2012pomdp,khouzaimi2017incremental,lopez2016automatic,su2016continuously,fatemi2016policy,asri2016sequence}.
The approaches here differ based on how the simulator itself is created or estimated, and whether or not the simulator is also considered an agent, which is trying to optimize its own reward.
For example, \citet{levin2000stochastic} tackle the problem of building a flight booking dialogue system.
They estimate a user simulator model by counting transition probabilities between dialogue states and user actions (similar to an n-gram model), which is then used to train a reinforcement learning policy.
In their setting, the states and actions are all abstract discrete variables, which minimizes the amount of natural language understanding and generation the policy has to learn.
As another example, \citet{georgila2011reinforcement} tackle the problem of learning dialogue policies for negotiation games, where each party in the dialogue is an agent with its own reward function.
In their setting, each policy is in effect also a user simulator, and is trained by playing against other policies using model-free on-policy reinforcement learning.
As a more recent example, \citet{yu2016strategy} build a open-domain, chitchat dialogue system using reinforcement learning.
In particular, \citet{yu2016strategy} propose to learn a dialogue manager policy through model-free off-policy reinforcement learning based on simulations with the template-based system A.L.I.C.E.~\citep{wallace2009anatomy} with a reward function learned from crowdsourced annotations. This is shown to yield substantial improvements w.r.t.\@ both the overall appropriateness of each system response and the conversational depth of the dialogues (e.g.\@ how long the system remains on topic).

Researchers have also recently started to investigate learning generative neural network policies operating directing on raw text through user simulations~\citep{li2016deep,das2017learning,lewis2017dealornodeal,liu2017iterative,lewis2017dealornodeal}.
In contrast to earlier work, these policies require both a deeper understanding of natural language and an ability to generate natural language.
For example, \citet{li2016deep} propose to train a generative sequence-to-sequence neural network using maximum log-likelihood, and then fine-tune it with a multi-objective function. The multi-objective function includes, among other things, a reinforcement learning signal based on self-play Monte Carlo rollouts (i.e.\@ simulated trajectories are generated by sampling from the model, similar to~\citep{silver2016mastering}) using a hand-crafted reward function.
\citet{lewis2017dealornodeal} apply model-free reinforcement learning for learning a system capable of negotiation in a toy domain from crowdsourced data. They demonstrate that it's feasible to learn an effective policy by training a generative sequence-to-sequence neural network on crowdsourced data, and that the policy can be further improved using on-policy reinforcement learning through self-play and Monte Carlo rollouts.
Both \citet{li2016deep} and \citet{lewis2017dealornodeal} use self-play.
Self-play is a viable option for training their policies because their problems are symmetric in the policy space (e.g.\@ any policy performing well on one side of the negotiation game will also perform well on the other side).
In contrast, self-play is unlikely to be an effective training method in our case, because the interactions are highly asymmetric: human users speak differently to our system than they would to humans and, further, they expect different answers.
\citet{liu2017iterative} use model-free on-policy reinforcement learning to improve a system in a restaurant booking toy domain. For training the system policy, they employ a user simulator trained on real-world human-human dialogues. In particular, under the constraint that both the system and the user share the exact same reward function, they demonstrate that reinforcement learning can be used to improve both the system policy and the user simulator. 
In a related vein, \citet{zhao2016towards} learn an end-to-end neural network system for playing a quiz game using off-policy reinforcement learning, where the environment is a game simulator. They demonstrate that combining reinforcement learning with dialogue state tracking labels yields superior performance.

In all the work reviewed so far, user simulators have been defined as rule-based models (e.g.\@ A.L.I.C.E.\@), parametric models (e.g.\@ n-gram models, generative neural networks), or a combination of the two.
In most cases, given a user simulator, the collected training data is discarded and the policy is learned directly from simulations with the user simulator.
In contrast, the Abstract Discourse MDP that we propose is a non-parametric approach, which repeatedly uses the collected training data during policy training.


Reinforcement learning has also been applied to teaching agents to communicate with each other in multi-agent environments \citep{foerster2016learning,sukhbaatar2016learning,lazaridou2016towards,lazaridou2016multi,mordatch2017emergence}.




\section{Future Work}
\subsection{Personalization}
One important direction for future research is personalization, i.e.\@ building a model of each user's personality, opinions and interests.
This will allow the system to provide a better user experience by adapting the response models to known attributes of the user.
We are in the process of implementing a state machine that given a user id, retrieves the relevant information attributes of the user from a database. 
If a particular user attribute is missing, then the state machine will ask the user for the relevant information and store it in the database.
One important user attribute is the user's name.
If no name is found in the database, the state machine may ask the user what they would like to be called and afterwards extracts the name from the user's response.
If a personal name is detected, it is stored in the database to be available for other modules to insert into their responses.
Name detection proceeds as follows.
First we match the response against a small collection of templates, such as "my name is ..." or "call me ...". 
Then we use part-of-speech (POS) tags of the resulting matches to detect the end boundary of the name. To avoid clipping the name too early due to wrong POS tags, we also match words against a list of common names in the 1990 US Census data\footnote{Obtained from: \url{https://deron.meranda.us/data/}.}.

In the future, we plan to explore learning user embeddings from previous interactions with each user, since we know from previous experiments that text information alone contains a significant amount of information about the speaker's identity~\citep{serban2015text}. Learning an embedding for each user will allow the system to become more personalized, by providing our response models with additional context beyond the immediate dialogue history.

\subsection{Text-based Evaluation}: It is well known that speech recognition errors have a significant impact on the user experience in dialogue systems~\citep{raux2006doing}.
Furthermore, speech recognition errors are likely to have a particularly averse effect on our system, because our system encourages open-ended, unrestricted conversations. Unlike many goal-driven and rule-based systems, our system does not \textit{take control} of the dialogue or direct the user to respond with a keyword from a set of canned responses.\footnote{In contrast, one socialbot system in the Alexa semi-finals would start the conversation by asking the user a question such as \textit{"I am able to talk about news, sports and politics. Which would you like to talk about?"} after which the user is expected to mention one of the keywords \textit{"news"}, \textit{"sports"} or \textit{"politics"}. This type of \textit{system-initiative} greatly reduces the number of speech recognition errors, because it is far easier to discriminate between a few keywords compared to transcribing a complete open-ended utterance.}
Because the users are more likely to give open-ended responses, the system is also more likely to suffer from speech recognition errors.
As we discussed in Section \ref{sec:model_selection_policy}, we did indeed observe a negative correlation between the confidences of the speech recognition system and the Alexa user scores.
Moreover, it is likely that speech recognition errors have a stronger systematic effect on some of the policies evaluated in Section \ref{sec:ab_testing_experiments}.

To mitigate the issues of speech recognition errors, we plan to evaluate the system with different policies through a text-based evaluation on Amazon Mechanical Turk. This would also help reduce other problems, such as errors due to incorrect turn-taking (e.g.\@ when the system barges in on the user, who is still speaking)~\citep{ward2005root}.



\section{Conclusion}
We have proposed a new large-scale ensemble-based dialogue system framework for the Amazon Alexa Prize competition.
Our system leverages a variety of machine learning techniques, including deep learning and reinforcement learning.
We have developed a new set of deep learning models for natural language retrieval and generation, including recurrent neural networks, sequence-to-sequence models and latent variable models.
In addition, we have developed a novel reinforcement learning procedure and evaluated it against existing reinforcement learning methods in A/B testing experiments with real-world users.
These innovations have enabled us to make substantial
improvements upon our baseline system.
On a scale $1-5$, our best performing system reached an average user score of $3.15$, with a minimal amount of hand-crafted states and rules and without engaging in \textit{non-conversational activities} (such as playing games or quizzes).
The performance is substantially above the average of all teams in the competition semi-finals, which was only $2.92$.
Furthermore, the same system averaged a high $14.5-16.0$ turns per conversation, which is substantially above both the average of all teams and the average of finalist teams in the competition semi-finals, suggesting that our system is one of the most \textit{engaging} systems in the competition.
Since nearly all our system components are trainable machine learning models, the system is likely to improve greatly with more interactions and additional data.

\subsubsection*{Acknowledgments}
We thank Aaron Courville, Michael Noseworthy, Nicolas Angelard-Gontier, Ryan Lowe, Prasanna Parthasarathi and Peter Henderson for helpful advice related to the system architecture, crowdsourcing and reinforcement learning throughout the Alexa Prize competition.
We thank Christian Droulers for building the graphical user interface for text-based chat.
We thank Amazon for providing Tesla K80 GPUs through the Amazon Web Services platform.
Some of the Titan X GPUs used for this research were donated by the NVIDIA Corporation.
The  authors  acknowledge  NSERC,  Canada  Research  Chairs,  CIFAR,  IBM  Research,  Nuance Foundation, Microsoft Maluuba and Druide Informatique Inc. for funding.


\bibliographystyle{agsm}
\bibliography{papers}

\end{document}